\documentclass[sigconf]{acmart}

\usepackage{booktabs} % For formal tables

\setcopyright{rightsretained}
\usepackage{amsmath,amssymb,amsfonts,bm,amsthm}
\usepackage{booktabs} % For formal tables
\usepackage{booktabs}
\usepackage{natbib}

\usepackage{graphicx}  %Required
\usepackage{amsmath}
\usepackage{bm}
\usepackage{verbatim}
\usepackage{subfigure}
\usepackage{amsfonts}
\usepackage[noend]{algpseudocode}

\usepackage{algorithmicx,algorithm}
\newtheorem{Definition}{Definition}
\newtheorem{Assumption}{Assumption}
\newtheorem{Theorem}{Theorem}
\newtheorem{Lemma}{Lemma}
\newtheorem{propos}{Proposition}

\begin{document}
	\title{Res-embedding for Deep Learning Based Click-Through Rate Prediction Modeling}
	\author{Guorui Zhou, Kailun Wu, Weijie Bian, Zhao Yang, Xiaoqiang Zhu, Kun Gai}
	%\authornote{Q. Pi and W. Bian share the co-first authorship.} 
	\affiliation{%
  		\institution{Alibaba Group}
	}
%\affiliation{$^\dag$Alibaba Inc.}
	\email{{guorui.xgr, kailun.wukailun, weijie.bwj, haoyi.yz, xiaoqiang.zxq, jingshi.gk}@alibaba-inc.com}

\begin{abstract}
Recently, click-through rate (CTR) prediction models have evolved from shallow methods to deep neural networks. Most deep CTR models follow an Embedding\&MLP paradigm, that is, first mapping discrete id features, e.g. user visited items, into low dimensional vectors with an embedding module, then learn a multi-layer perception (MLP) to fit the target. In this way, embedding module performs as the representative learning and plays a key role in the model performance. However, in many real-world applications, deep CTR model often suffers from poor generalization performance, which is mostly due to the learning of embedding parameters. In this paper, we model user behavior using an interest delay model, study carefully the embedding mechanism, and obtain two important results: (i) We theoretically prove that small aggregation radius of embedding vectors of items which belongs to a same user interest domain will result in good generalization performance of deep CTR model. (ii) Following our theoretical analysis, we design a new embedding structure named res-embedding. In res-embedding module, embedding vector of each item is the sum of two components: (i) a central embedding vector calculated from an item-based interest graph (ii) a residual embedding vector with its scale to be relatively small. Empirical evaluation on several public datasets demonstrates the effectiveness of the proposed res-embedding structure, which brings significant improvement on the model performance.

	\end{abstract}

	\maketitle
	\section{Introduction}

%In the advertising and recommendation systems, CTR prediction is a crucial task to decide final ranking of items presented to the consumers. It receives features of a user and the features of the target item as inputs and outputs the probability of this user clicking on the target item under the e-commerce scenario. The CTR prediction tasks has attracted a lot of attention and become the cornerstone of advertising and recommendation systems in many large enterprises.

In recommender systems, CTR (click-through rate) prediction is a crucial task, which has attracted a lot of attention and become the cornerstone. The aim of CTR model is to predict the probability of one user click a given candidate item, which will decide the final ranking of items presented to users.

\begin{figure}[t]
	\centering
	\includegraphics[scale=0.27]{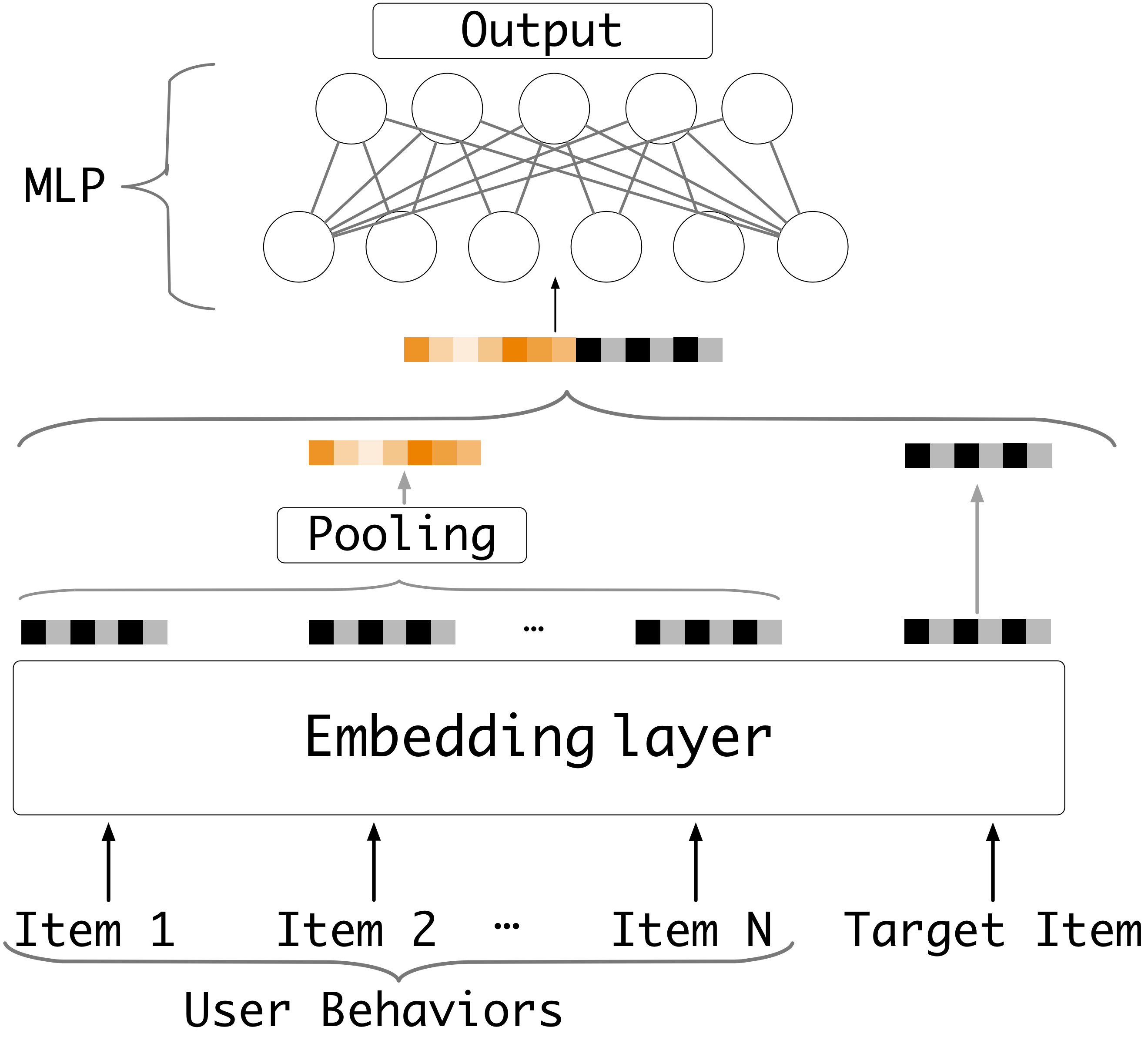}
	\caption{The illustration of the Embedding\&MLP structure.}
	\label{fig:mlpemb}
	%\vspace{-0.6cm}
\end{figure}

%Thanks to the rapid development of deep learning technology\cite{lecun2015deep} \cite{krizhevsky2012imagenet} \cite{he2016deep} \cite{bahdanau2014neural},
Thanks to the rapid development of deep learning technology\cite{lecun2015deep}, deep neural network based models for CTR prediction task have gained significant progress and become state-of-the-art methods. Most deep CTR models follow a Embedding\&MLP paradigm, as illustrated in Fig.\ref{fig:mlpemb}: (i) \textbf{Embedding} module which maps discrete id features, e.g., user historical clicked items, into low dimensional vectors and then transformed into a fixed-length vector by pooling. (ii) \textbf{MLP} module which aims to learn the nonlinear relationship among features and fit the target by a fully connected network, a.k.a. multi-layer perception. 

Many approaches have been proposed to improve the performance of deep CTR model, such as PNN\cite{Qu2017Product}, DeepFM\cite{Guo2017DeepFM}, DIN\cite{Zhou2017Deep} and DIEN\cite{Zhou2018Dien}. However, most work focus on designing new network architectures to substitute MLP module, aiming to better capture the nonlinear relationship among features, while little effort has been made to improve the basic yet important embedding module. In this paper, we study carefully the embedding mechanism theoretically and try to fill the gap in practical.  
   
Generally speaking, traditional embedding module works in a look-up table manner, that is, each discrete feature corresponds to a low dimensional vector, with parameters learned from training data of CTR task. Note that parameters of both embedding module and MLP module are learned end-to-end. In this way, embedding module actually performs as a representative mapping and determines the input distribution of the subsequent MLP module. Referring to the data-dependent generalization theory\cite{Kawaguchi2018Generalization}, input distribution will influence the generalization performance of model. Therefore, embedding module is vital for the generalization performance of deep CTR model. As \cite{Zhou2017Deep} reported, in practice, overfitting phenomenon commonly exists during training of deep CTR model, especially in industrial applications with large scale discrete features. We argue that it is the embedding module that might cause the poor generalization performance. The reason lies in two-folds: (i) In many real systems, number of features can scale up to billions, causing the number of embedding parameters to be huge. This would promote the memory ability but decrease the generalization ability. (ii) With the supervision of click labels only, it might be hard for the traditional embedding module to learn a representative mapping with high generalization ability. For example, distance of embedding vectors of two similar items might change greatly with different initializations in the end-to-end training manner. 

Motivated by the above observation, in this paper, we propose to (i) quantitatively analyze which variables are involved in the generalization error bound of deep CTR models, and (ii) design corresponding solutions to enhance generalization ability according to this quantitative relationship. 

We take the recommender system in e-commerce industry as an example. In e-commerce scenarios, we shall first model users' behavior with an “interest-delay” model. According to experience, we suppose that user's interests will last for a while and users' click behaviors are generally affected by their interests. Different interests will drive users to click different types of items. Hence, assume that each item has its own interest domain. E.g., item "iphone 6" might belong to the interest domain of "smartphone". Intuitively, items belonging to the same interest domain should be similar and distances between their embedding vectors should be small. In fact, we do have proved this mathematically. Specifically, \textbf{we prove that the generalization error of deep CTR model is bounded by the envelope radius of items with the same interest domain in the embedding space}. Moreover, following this theoretical analysis, we design a new residual embedding structure which is theoretically helpful for improving the generalization ability named as res-embedding. In this structure, \textbf{embedding vector of each item is the sum of a central embedding vector and an independent low-scale residual embedding vector}. Items in the same interest domain have similar central embedding vectors. To achieve this goal, we build an item-based interest graph based on the co-occurrence frequency of items in user historical behavior sequences. Central embedding vector of each item is calculated as the linear combination of the central embedding basis vectors of its neighboring items in the interest graph, with the linear combination coefficients calculated by three practical implementations including average, GCN (Graph Convolutional Network)\cite{Kipf2016Semi} and attention. Besides, the residual embedding vector is forced to be small-scale, by penalizing the $l_2$-norm of parameters of residual embedding vectors on the final objective function of deep CTR model.

Contributions of this paper are summarized as follows:
\begin{itemize}
	\item We theoretically proved that increasing the aggregation degree of embedding vectors of items in the same interest domain helps decrease the generalization error bound. This may be a direction worth studying for the future to improve the generalization performance of embeddings.
	\item Following the theoretical analysis, we propose a new res-embedding structure as well as three practical implementations including average, GCN and attention. In addition, We also mathematically prove the consistency of theory and method.
	\item We conduct careful experiments on several public datasets. Adding with res-embedding structure, state-of-the-art deep CTR models all gains significant improvement on the AUC metric. This clearly verifies the proposed theory as well as res-embedding method.      
\end{itemize}

The rest of the paper is organized as follows. The related work is summarized in section 2. Section 3 theoretically analyzes the influence of the distribution of embedding vectors on the generalization performance of deep CTR prediction model.Then we propose the detail of the res-embedding and explain its consistency with theoretical analysis in section 4. Our experimental results are shown in section 5 and conclusion is shown in section 6.\vspace{-0.3cm}

	\section{Related Work}

Deep CTR model: Recently, the deep learning has been widely used in the CTR prediction task. In the very first, NNLM \cite{Bengio2003A} learns the representation of each word, which has inspired many natural language models and CTR prediction models that need to handle large-scale sparse input features. Piece-wise Linear Models (LS-PLM) \cite{Gai2017Learning} and factorization machine (FM) \cite{Rendle2011Factorization} adopt the embedding layer for the sparse input feature and capture the relationship amongs the different features through the specific form functions, which can be regard as a single-layer neural network. Based on these model, many deeper models like Deep Crossing \cite{Shan2016Deep}, Wide\&Deep Learning \cite{Cheng2016Wide} and YouTube Recommendation CTR model \cite{Covington2016Deep} employ more complex MLP. PNN\cite{Qu2017Product} tries to capture high-order feature interactions by involving a product layer after embedding layer. Therefore, there are also some works\cite{Zhou2017Deep} that extract advanced information with more abstract features. Though improvements based on deep model continuously enhance the performance of CTR prediction tasks, their inputs are mostly from the original embedding layer. We design embedding layer to promote the generalization performance based on these networks.

Embedding learning: Traditional methods calculate embedding vector by the relationship between high dimensional data. Some embedding algorithm like Laplacian Eigenmaps \cite{Belkin2002Laplacian} and LLE (Locally Linear Embedding) Graph Factorization \cite{Roweis2000Nonlinear}, LINE \cite{Tang2015LINE}  keep stronger related nodes closer to each other in the vector space. Some deep methods like GCN \cite{Kipf2016Semi}  define a convolution operator on graph to learn the embedding based on the graph. The model iteratively aggregates the embeddings of neighbors for a node and uses them to obtain the new embedding. Individual literature \cite{Parsana2018Improving} uses an additional network and context information of the data to pre-train embedding vectors to predict the CTR model.

Generalization: \cite{Xu2012Robustness} proposes a theoretical framework to define the robustness of learning algorithm and proves that generalization performance of the learning algorithm is determined by the robustness of a learning algorithm. It provides the robustness of some common learning algorithms like SVM, DNN and so on. This theoretical framework and example based on DNN help us to theoretically analyze the influence of distribution of embedding vectors on the generalization of deep CTR model. \cite{zantedeschi2016metric} utilizes this framework to study the relationship of metric structure of the features and robustness of the algorithm and proposes structured metric learning method. Other similar work \cite{trofimov2017representation}\cite{usunier2006generalization} discuss the structure of the embedding should be tide. Compared with these work, our paper study more complex structure of embedding vectors. With modeling based on the interest state, we prove that embeddings under a kind of group aggregated structure could promote the generalization of the deep model. 
	\section{Theoretical Analysis}
\label{Analysis}
In this section, we shall first model the user's click behavior above the user's interest called interest delay model. For ease of expression, we divide interest into several categories, called interest domains. Then, the mathematical discussion of generalization of CTR model is based on the interest delay model. According to some generalization theory, we finally come to the conclusion that the generalized error bound can be effectively reduced by reducing the envelope radius of the items belonging to the same interest domain in the embedding space without greatly changing the total envelope radius of all items in the embedding space. Based on this conclusion, we propose a basic prototype of res-embedding and its final improved version.

\subsection{Notation}
\begin{table}[h]
	\label{table_1}
	\begin{tabular}{|p{2.5cm}|p{5cm}|}
		\hline
		Expression&Meaning\\
		\hline
	$d$& The dimension of embeding space \\
			\hline
$\{x^1,\cdots,x^M\}_h$&The user's click behavior sequence with length\\
		\hline
$x_t$&  The target item \\
		\hline
$y^*$&  Ground-truth label\\
		\hline
$f(\cdot,\cdot)$&  Deep CTR model\\
		\hline
$z$&  Interest hidden state\\
		\hline
$N_z$&  Number of the Interest hidden state\\
		\hline
$P(x|z)$&  Conditional distribution of item $x$ under the Interest hidden state $x$\\
		\hline
$P(x),P(z)$ &The marginal distribution of item $x$ and prior distribution of interest hidden state $z$  \\
		\hline
$T,p$&  Number of the preriods and the time step\\
		\hline
$N$&  The number of the training samples\\
		\hline
$ l(f,s)$&  Loss function of the sample $s$ based on the model $f$\\
		\hline
$E_s(l(f,s))$&  Expected loss \\
		\hline
$\sum_{i=1}^{N}l(f,s_i))/N$& Empirical loss \\
		\hline
$\phi(V)$&  Envelope radius of the set $V$\\
		\hline
	\end{tabular}
	\caption{The notations of some main Expressions and their meanings.}
\end{table}
The expressions and variables in Table \ref{table_1} are some important mathematical symbols in this section. The specific definition and detailed introduction are shown as follows.

\textbf{Embedding Space}: In this section, all of the items have their own representation vectors in the $d$-dimensional embedding space $\mathbb{R}^d$; We mainly analyze the influence of the distribution of items in this embedding space on CTR model. For simplicity, item with embedding vector is referred to as item itself in this section (e.g. "item $x$" means item with embedding vector $x$.)

\textbf{Data and model}: In CTR prediction task, one sample $s$ is compose of the input features and label. That is $s = \{\{\{x^1,\cdots,x^M\}_h,x_t\},y^*\}$. The input features of one sample are composed of user's behavior sequence, i.e. the clicked item sequence $\{x^1,\cdots,x^M\}_h$, ($x^i\in \mathbb{R}^d$ means item clicked in $i$-th time step, $M$ is the total number of time steps) and the target item $x_t \in \mathbb{R}^d$. The label of the sample $y^* \in \{0,1\}$ indicates whether the user clicked on the target item. The CTR prediction model $f(\cdot,\cdot):(\mathbb{R}^{M\times d},\mathbb{R}^{d})\rightarrow [0,1]$ is learned by fitting training samples. In this paper, we shall discuss the paper around the $D$ layers MLP with ReLU nonlinear function which is defined in Definition \ref{MLP_Relu}

\begin{Definition}
	\label{MLP_Relu}
	$D$ layers MLP with ReLU nonlinear function and $n\times d$ input dimensions is defined as follows. The input is  concatenation of $\{x_1,\cdots,x_n\}$ and $x_i \in \mathbb{R}^d$ for $\forall i$, and the label is $y^*\in \mathcal{Y} (\mathcal{Y}=\{0,1\})$. The trainable parameters are $\{W_t,b_t\}$ for $\forall t = 1,\cdots,D$. Its forward process is
	\begin{equation}
	\begin{aligned}
	&h^0=[x_1,\cdots,x_n],\\
	&h^t = ReLU(W_th^{t-1}+b_t),\\
	&f(x_1,x_2,\cdots,x_n) = sigmoid(W_Dh^{D-1}+b_{D}).
	\end{aligned}
	\end{equation}
	$ReLU(x)=max(x,0)$ and the $sigmoid(x)$ is the sigmoid function. 
	
\end{Definition}

\textbf{Generalization error bound}: Generalization error bound of model $f$ is the upper bound of the absolute value of the difference between expected loss $E_s(l(f,s))$ and empirical loss $\sum_{i=1}^{N}l(f,s_i)/N$. $l(f,s)$ is the loss function of the ground-truth label of $s$ and the output of $f$ when the input is the feature of $s$. Empirical loss is the expectation of loss based on distribution of sample $s$,i.e $E_s(l(f,s))=\int_{s}l(f,s)p(s)$, and experience loss is the average loss of all training samples $s_i$, $i\in \{1,\cdots,N\}$. $N$ is the number of training samples.

\textbf{Envelope radius}: For a set $V \subset \mathbb{R}^d$, if $\exists w \in \mathbb{R}^d $ s.t. $\forall s \in V$, $\|w-s\|_2\leq R_0$, then envelope radius of set $V$ is no more than $R_0$, which is defined as $\phi(V)\leq R_0$. 

\subsection{User Behavior Modeling: Interest Delay Model}
\label{IDM}
  \begin{figure}[ht]
	\setlength{\abovecaptionskip}{-0.03cm}
	\centering
	\includegraphics[scale=0.3]{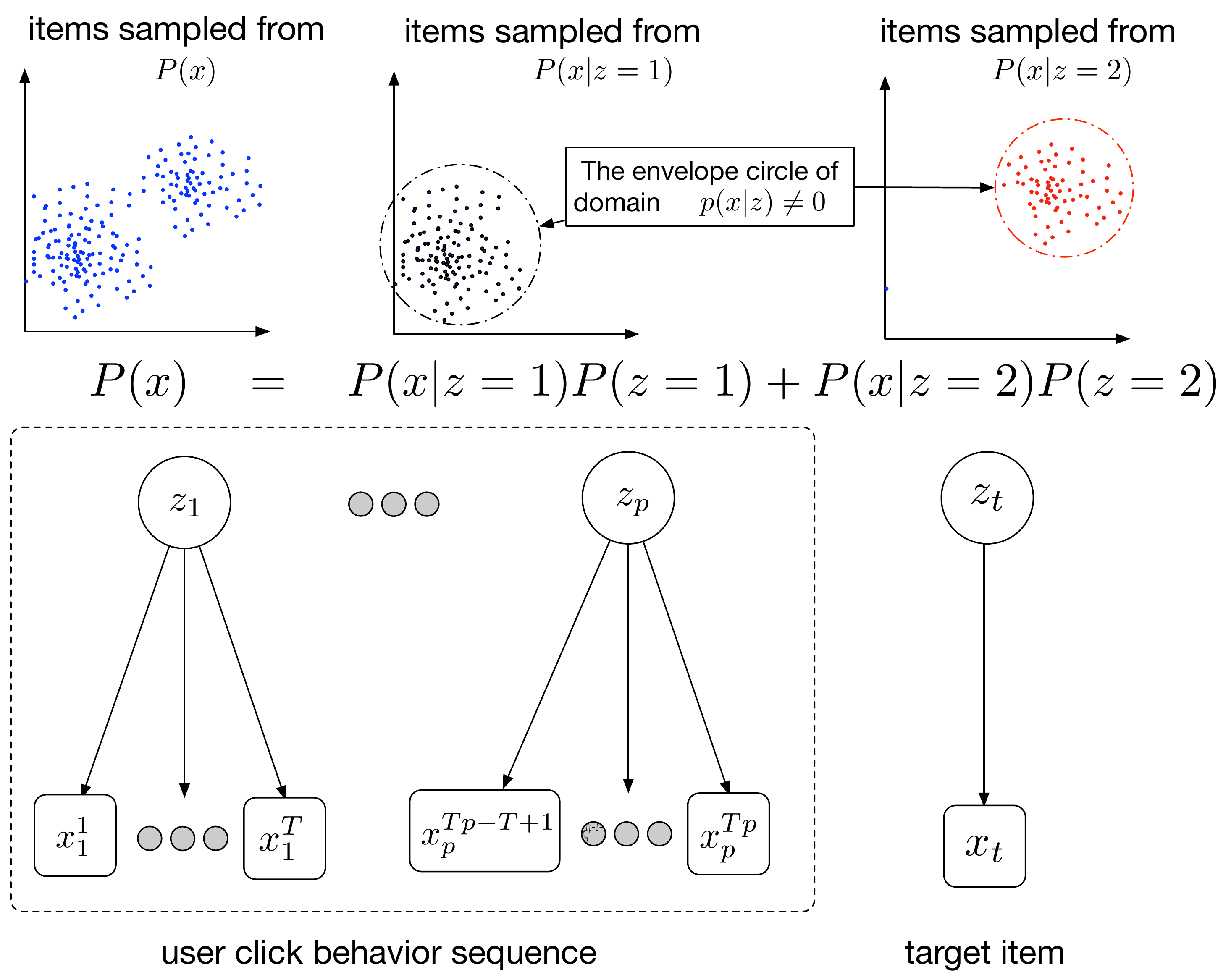}
	\caption{The upper part of this figure is the case of $N_z=2$. The distribution of item embedding $P(x)$ is composed of conditional distribution $P(x|z=1)$ and $P(x|z=2)$. The two dashed circle is the envelope circle of the domain$P(x|z)\neq 0$. The lower part of this figure is the sampling process of behavior sequence and target item of one sample. The behavior sequence is generated from the hidden state of $p$ time periods from $z_1$ to $z_p$. One hidden state controls $T$ time steps.}
	\label{fig:z}
	\vspace{-0.2cm}
\end{figure}

Intuitively, when browsing products in a E-commerce website, users will click different products due to different interest. Based on experience and intuitive common sense, we assume that the user's interest will last for a period of time when browsing the internet. Each click is defined as one “time step” and the sequence of the time steps with the same interest is one "period". This model is so called "interest delay" model.

Focusing on a single time step, the user's single click behavior is determined by the user's current interests. We mathematically illustrate this behavior as the sampling process as follows. Assume that there are $N_z$ interest domains, and interest hidden state $z \in \{1,2,\cdots,N_z\}$ is defined as a quantitative discrete variable of interest domain. The probability distribution function $P(x)$ of one user click item $x$ is related to his interest hidden state $z$ as $P(x|z)$. The distribution of items in embedding space consists of conditional distribution $P(x) = \sum_{z}P (x | z) P (z)$, where $P (z)$ is the prior probability distribution of $z$. As shown in upper part of Figure \ref{fig:z}, the total distribution of the clicked item $P(x)$ can be regarded as a combination of two conditional distributions $P(x|z=1)$ and $P(x|z=2)$. By the way, we define the domain of $P(x|z)$ is the area that $P(x_0|z=i)\neq 0$. 

A click behavior sequence of user in interest delay model is generated from two phases: Firstly, a user determine an interest hidden state $z_i$ in each period(Totally $p$ periods) with $T$ time steps, and then sample (randomly click) a item $x_{p_i}^t$ with the influence of interest hidden state $z_i$. Thus, user's clicked items sequence could be written as $\{\{{x_{p_1}^1},{x_{p_1}^2},\cdots,{x_{p_1}^T}\},\{{x_{p_2}^{T+1}},\cdots,{x_{p_2}^{2T}}\}$ $,\cdots,\{{\cdots,x_{p_p}^{T\times p}\}}\}_h$ ($p_i$ means the $i$-th period, $M=T\times p$). The sampling process of behavior sequence is shown in lower part of Figure \ref{fig:z}. Assumption \ref{Assum2} summarizes the quantitative sampling process of user behavior in the interest delay model.

 %The interest hidden state $z$ is the most important definition of this paper. Through setting of this hidden state, the marginal distribution of the items $P(x)$ is reconstructed to several conditional distribution $P(x)=\sum_{i}P(x|z=i)P(z=i)$. We can describe the embedding structure using the conditional distribution $P(x|z=i)$, which is much more refined than using the marginal distribution to describe the embedding structure. 

\begin{Assumption}
	\label{Assum2}
	Training sample 
	
	$s = (\{\{{x_{p_1}^1},\cdots,{x_{p_1}^T}\},\{{x_{p_2}^{T+1}},\cdots,{x_{p_2}^{T\times 2}}\},\cdots
	,\{x_{p_p}^{T\times(p-1)+1},\cdots,$ $x_{p_p}^{T\times p}\}\}_h,x_{t},y^*)$
	consists of clicked items $\{x^1,\cdots,x^{T\times p}\}_h$ in $p$ periods with $T$ time steps, target item $x_t$ and label $y^*$. The interest hidden state of target item $z_t$ is sampled from $P_t(z)$, and target item $x_t$ are sampled from the conditional distribution $P(x|z)$. For user's click behavior,
	interest hidden state sequence of $p$ periods $\tilde{z}=\{z_1,\cdots,z_p\}$ is sampled from a set $\tilde{S}_z\subset \bigcup_1^{p}\{1,\cdots,N_z\}$. Each element of $\tilde{z}$ controls the user's click behavior in $T$ time steps. 
	That is, each item of historical click behaviors subsequence $\{x^{T\times(i-1)+1},\cdots,x^{T\times i}\}_h$ 
	is sampled from the conditional distribution $P(x|z_i)$. $z_i$ is interest hidden state in the $i$-th period. The number of elements in set $\tilde{S}_z$ is $N_{S}$. The label $y^*$ is sampled from $\{0,1\}$.
\end{Assumption}
The sampling process of behavior sequence is shown in Figure \ref{fig:z}.
 It assumes that one interest hidden state can control a $T$ time steps peroid, which is consistent with the actual situation for the user's interest always lasts for a while in the e-commerce scenario. Interest hidden of click behavior is a sequence $\tilde{z}$ of $p$ periods, which is sample from a set $\tilde{S}_z$. Therefore, the final length of click behavior sequence length should be $T\times p$.

\subsection{Analysis of Generalization Error Bound on Interest Delay Model}
Section \ref{IDM} models user behavior quantitatively through the interest delay model. In this subsection we will discuss generalization error of the DNNs on the data generated by the interest delay model. Which variable is related to generalization error? Theorem \ref{Theorem2} will give a generalization error bound of the MLP with ReLU in definition \ref{MLP_Relu} on the data sampled as the setting of Assumption \ref{Assum2}.

\begin{Theorem}
	\label{Theorem2}
	
		The data sample represented as $s=\{\{x^1,\cdots,x^{T\times p}\}_h,x_t,y^*\}$ is generated from the way of Assumption \ref{Assum2}. For the $D$ layers MLP with ReLU and $(Tp+1)\times d$ input dimensions defined in Definition \ref{MLP_Relu}, $f(\{x^1,\cdots,x^{T\times p}\}_h,x_t)$, the loss function of MLP is $f(l,s) = |f(\{x^1,\cdots,x^{T\times p}\}_h,,x_t)-y^*|$. If there are $N$ samples sampled from the way of Assumption \ref{Assum2} to train the MLP, with the probability $1-\delta$, the generalization error bound satisfies
\begin{equation}
\label{equation_of_theorem2}
\small{
\begin{aligned}
&|E_s(l(f,s))-\frac{1}{N}\sum_{i=1}^{N}l(f,s_i)| \leq \inf_r\{\sqrt{Tp+1}\|W\|_2^Dr\\
&+ l_M\sqrt{\frac{4N_zN_S(2R_{max}\sqrt{d}/r)^{d(Tp+1)}ln2+2ln(1/\delta)}{N}}\}
\end{aligned}}
\end{equation}

	Among them, $r$ is a parameter and will disappear in the infimum operation. $\|W\|_2$ is the average of 2-norm of all parameter matrices in MLP. That is $\|W\|_2 = \sum_{i=1}^{D}\|W_i\|_2/D$. $l_M$ is the maximum value of $l(f,s)$, and $l_M=1$ in Definition \ref{MLP_Relu}. $N_z$ is the number of the interest domains and $N_S$ is the number of the interest sequence set. $R_{max} = \max_i\phi(domain P_i)$ for all $i=1,\cdots,N_z$. $d$ is dimension of embedding vector.
\end{Theorem}

As long as the structrue of model and the training data are fixed, $D$, $d$, $p$, $T$, $N$, $N_z$ and $N_S$ are fixed too. According to Theorem \ref{Theorem2}, \textbf{generalization error will only be affected by the $\|W\|_2$ and $R_{max}$}. We will not discuss $\|W\|_2$ here for $\|W\|_2$ is affected by too many factors that it is difficult to analyze the relationship between $\|W\|_2$ and embedding layer. Reducing the whole scale of the embedding vectors seems to corroborate the effectiveness of applying a regularization term of embedding layer in some cases. However, it will also reduce the capacity and representation ability of embeddings. By individually reducing the radius of envelope circles of each interest domain, one could \textbf{make the items with the same interest domain closer in embedding space but maintain the distances among the items of different interest domains}, which can control the generalization error bound and maintain capacity and representation ability of CTR prediction model at the same time.

The proof of the Theorem \ref{Theorem2} and the details of specific derivation process is in the supplementary material.

\subsection{Prototype and discussion}
\label{discussion}
Based on the theoretical analysis, we firstly propose and concentrate on a basic prototype. In this prototype, items in the same interest domain shared the same \textbf{central} embedding vectors and each item has its unique \textbf{residual} embedding vector with smaller scale. The final embedding vector of one item is the sum of its \textbf{central} and \textbf{residual} embedding vectors.
By reducing the scale of the \textbf{residual} embedding, we can effectively reduce the distance of the items of the same interest domain in embedding space, which will reduce $R_{max}$ in (\ref{equation_of_theorem2}).

 Assume there are $I$ interest domains and $H$ items totally. We define central embedding matrix as $\bm{C} \in \mathbb{R}^{I\times d}$. Each row of $\bm{C}$ is the central embedding vector of each interest domain. $\bm{P}\in \mathbb{R}^{H\times I}$ is the relationship matrix between items and interest domains. Each row of $\bm{P}$ is one-hot vector. If $i$-th item belongs to $j$-th interest domain, the element $P(i,j)=1$ otherwise $0$. $\bm{R}\in \mathbb{R}^{H\times d}$ is residual embedding matrix. Each row of $\bm{R}$ is the residual embedding vector of each item. The final embedding matrix $\bm{E} \in \mathbb{R}^{H\times d}$, and each row of $\bm{E}$ is the final embedding vector of each item. $\mathbb{E}$ is calculated as
 
\vspace{-0.05cm}
\begin{equation}
\bm{E} = \bm{P}\bm{C} + \bm{R}.
\vspace{-0.05cm}
\end{equation}

\begin{figure}[ht]
	\setlength{\abovecaptionskip}{-0.03cm}
	\centering
	\includegraphics[scale=0.30]{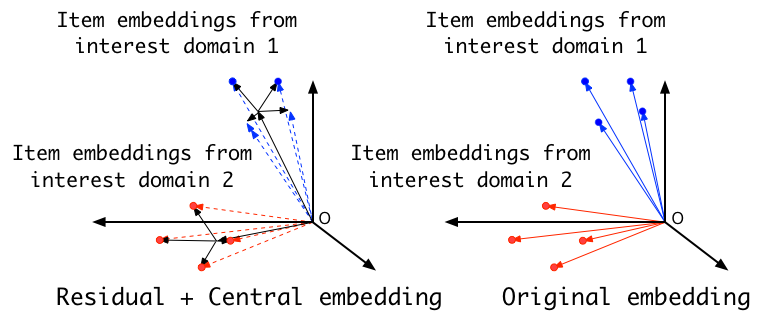}
	\caption{Sketch Map of the basic res-embedding prototype with $8$ items and $2$ interest domains ($H=8$ and $I = 2$) in the 3-d embedding space.}
	\label{fig:1}
	\vspace{-0.2cm}
\end{figure}

For instance, in Figure \ref{fig:1}, 8 embedding vectors are replaced into 2 central embedding vectors and 8 residual embedding vectors. With this structure, the envelope radius of embedding vectors in the same interest domain is bounded by the scale of residual embedding vectors.

However, the relationship matrix $\bm{P}$ is unknown and hard to obtain. In another word, it is unknown that which interest domain each item belongs to. Wrong $\bm{P}$ may bring worse generalization to deep CTR prediction model. It is necessary to ascertain the reasonable relationship between items and interest domain. Moreover, It is very likely that a item does not belong to only one interest domain. That is, the constraint relationship of central embedding could be soft, which means the central embeddings of items in the same interest domain may not be exactly the same, but just more similar.

In order to solve this problem, we reexamine Assumption \ref{Assum2} and notice that one hidden interest state $z$ will keep $T$ time steps to affect user click behavior. Therefore, a conclusion that two items are more likely to be in the same interest domain if they appear more frequently in a short-term of user click behaviors can be deduce. From this conclusion, we define an item interest graph $\bm{Z}$ constructed by co-occurrence frequency of each item pair to describe similarity relation of interest domain among items.

 %Accordingly how to ascertain the which interest domain the goods belong to becomes a crucial question. Based on our Assumption \ref{Assum2}, the same interest hidden state controls the distribution of embedding vectors of clicked goods in short time $T$.  Our proposed structure aggregates the embedding vectors of goods frequently appearing in user historical behavior for a short time. 
\begin{comment}
According to Assumption \ref{Assum2} we construct two settings:
1. The embedding vectors of two goods are more likely to be in the same interest domain if they appear more frequently in a short-term of user historical behaviors. 

2. The larger the possibility of goods in the same interest domain, the stronger the correlation between their central embedding vectors.

Both settings are based on our theoretical analysis. Setting 1 is derived from assumption 2. An interest hidden state controls the click behavior of the user in a short-term period. This also means that the goods that users click on in a relatively short period are more likely to be in the same interest domain. The higher the frequency of common clicks, the greater the possibility of being in the same interest domain. Setting 2 is derived from Theorem 1 and Theorem 2. The goods embedding in the same domain of interest should be closer to each other. Thus, the central embedding vectors should be closer
\end{comment}

	\section{Method}
In the Section \ref{Analysis}, there is a conclusion that the reducing the distance of items with the same interest domain in the embedding space is helpful to promote the generalization of CTR prediction model. However, \textbf{as our discussion in sub-section \ref{discussion}}, in the real scenario, the original prototype proposed in Section \ref{Analysis} has the problem that it is impossible to determine which interest domain each item belongs to for the detail information of interest domain of the is unknown. In this section, based on the theoretical analysis and the prototype, we construct an interest graph describing the similarity of interest domains between items, and propose a improved res-embedding base on the interest graph. As the evolutionary version of the proposed prototype, res-embedding contain central embedding part and residual embedding part. In central embedding part, Central embedding vector of each item is calculated through its adjacent items in the interest graph. In residual embedding part, each item owns an independent embedding vector.

\subsection{Res-embedding}

%According to the Theorem \ref{Theorem1} and Theorem \ref{Theorem2}, we discover that the embedding vectors with group aggregation will improve the back-end deep CTR prediction model. The grouping aggregation embedding vectors are shown in Figure \ref{fig:1}. Eight goods belongs to 2 insterest domain. 4 blue embedding vectors belong to the domain 1 and the other 4 embedding vectors belong to the domain 2. The group aggregation of the embedding vectors is reflected in the fact that the blue embedding vectors are close to each other, and the red embedding vectors have the same characteristic. 
  We still assume that there are $H$ items totally. Res-embedding structure is parameterized by two trainable parameter matrices, central embedding basis matrix $\bm{C_b}\in \mathbb{R}^{H\times d}$ and the residual embedding matrix $\bm{R} \in \mathbb{R}^{H\times d}$. Each row vector of $\bm{C_b}$ is central embedding basis vector of corresponding item and Each row vector of $\bm{R}$ is residual embedding vector of corresponding item. The final embedding vector of one item is the sum of its central embedding vector and its residual embedding vector. The residual embedding vector of one item can directly read from $\bm{R}$ and central embedding vector of one item is derived from the linear combination of other items' central embedding basis vectors. We define the residual embedding matrix $\bm{W} \in \mathbb{R}^{H\times H}$ and the $i$-th row vector in $\bm{W}$ represents the linear combination coefficients of $i$-item about all items central embedding bases. The calculation process of final embedding matrix $\bm{E}$ is shown as follows
 \begin{equation}
 \begin{aligned}
 \bm{E} =\bm{W}\bm{C_b} + \bm{R}.
 \end{aligned}
 \end{equation}
 Each row vector of $\bm{E}$ is the embedding vector of corresponding item finally entered into the MLP.

 According to the theoretical analysis, a reasonable linear combination matrix $\bm{W}$ should make the central embedding of the items with the same interest domain closer. In order to calculate a reasonable linear combination matrix $\bm{W}$, we utilize the information of item interest graph mentioned above. The central embedding of item should be a linear combination of the center embedding basis of items connected with it in the item interest graph.
 %The final central embedding of each goods is the linear combination of the central embedding basis of its similar goods under interest domains. In order to determine the similarity relationship of goods under interest domains. 
 We construct a item interest graph with its adjacent matrix $\bm{Z}\in \mathbb{R}^{H\times H}$ based on the co-occurrence frequency of items on historical click behavior sequence of all users according to the inference of Assumption \ref{Assum2}. %which is that items frequently co-occur in a short period are more likely in the same interest domain. 
 The construction process of the graph is illustrated in Algorithm \ref{alg:graph}. In Algorithm \ref{alg:graph}, the $B=\{b_1,\cdots,b_N\}$ is the set of user click behavior sequences of all $N$ training samples, and $b_i=\{g_1^i,\dots,g_{m_i}^i\}$ is the user click behavior sequence of $i$-th sample with $m_i$ clicked items arranged in time sequence. $g_{j}^i$ means the ID of $j$-th clicked item in $b_i$. Item ID is the index of the adjacent matrix, $\{1,2,\cdots,H\}$. 
 
  The linear combination matrix $\bm{W}$ is decuded from adjacent matrix $\bm{Z}$ of item interest graph.
   \begin{equation}
  \begin{aligned}
  \label{wgz}
  \bm{W} = g(\bm{Z})
   \end{aligned}
  \end{equation}
 function $g():\mathbb{R}^{N\times N}\rightarrow \mathbb{R}^{N\times N}$ converts the adjcent matrix to a linear combination matrix with the same size.
 
%According to setting 1, the goods more frequently co-occurrence in a short-term of user historical behaviors products are more likely to be in the same interest domain.
\begin{algorithm}[t]
	\caption{Graph construction} %算法的名字
	\label{alg:graph}
	\hspace*{0.02in} {\bf Input:} %算法的输入， \hspace*{0.02in}用来控制位置，同时利用 \\ 进行换行
	user click behavior sequences set $ B=\{b_1,\cdots,b_n\}$, window radius $\Delta$\\
	\hspace*{0.02in} {\bf Output:} %算法的结果输出
	Adjacent matrix $\bm{Z}$
	\begin{algorithmic}[1]
		\State Initialize the all element of adjacent matrix $\bm{Z}$ as 0. % \State 后写一般语句
		\For{$i$ in $1 \cdots N$} 
		\State $b_i=\{g_1^i,\dots,g_{m_i}^i\} $% For 语句，需要和EndFor对应
		\For{$g_c^i$ in $\{g_1^i,\dots,g_{m_i}^i\}$}
		\State $ N_l = max(1,c-\Delta)$
		\State $ N_r = min(c+\Delta,m_i)$
		\For{$g_w$ in $\{g_{N_l}^i,g_{N_l+1}^i,\dots,g_{N_r}^i\}\backslash g_c$} 
		\State $\bm{Z}(g_c,g_w) \leftarrow \bm{Z}(g_c,g_w) +1$
		\EndFor
		\EndFor
		\EndFor
		\State Keeping the largest $K$ values in each row of $\bm{Z}$, and set the rest to 0
		\State \Return $\bm{Z}$
	\end{algorithmic}
\end{algorithm}
\begin{comment}
1. Initializing the connection graph. The Laplace matrix $\bm{A}$ is set as 0. 

2. Window traversing the good $(b_1, \cdots, b_n) $ in the user behavior sequence, centers goods $b_i$, and takes the length of $2W+ 1$ as the window size. All of elements the Laplace matrix weight of goods $(b_{i-W}, \cdots, b_{i-1}, \cdots, b_{i+1},  b_{i+W})$ and good $b_i$ will be added by 1.

3. Keeping the largest $K$ values in each row, and the rest values are set to 0 to generate the final graph connection matrix $\bm{L}$.

%After constructing a full graph of goods,  In the graph, the bigger the connection weight between goods A and goods B, the lager possibility of belonging to the same interest domain. According to our theory, the embedding vector of goods A should be closer to the embedding vector of goods B. If the embedding vector of goods B is added in the original embedding vector of goods A, we can achieve this goal. Therefore, when the center embedding vector of a good is expressed, in addition to its own central embedding vector, it also fuses the central embedding vector of adjacent goods according to certain mechanism. Specifically, as shown in Figure \ref{fig:2}, 
\end{comment}
The final central embedding of one item is composed of linear combination of the central embedding basis of its adjacent items in the interest graph $\bm{Z}$. The linear combination coefficients $w$ are determined by the connection weight of the graph.

\begin{figure}[!t]
  \setlength{\abovecaptionskip}{-0.03cm}
	\centering
	\includegraphics[height=2.2in, width=3.3in, keepaspectratio]{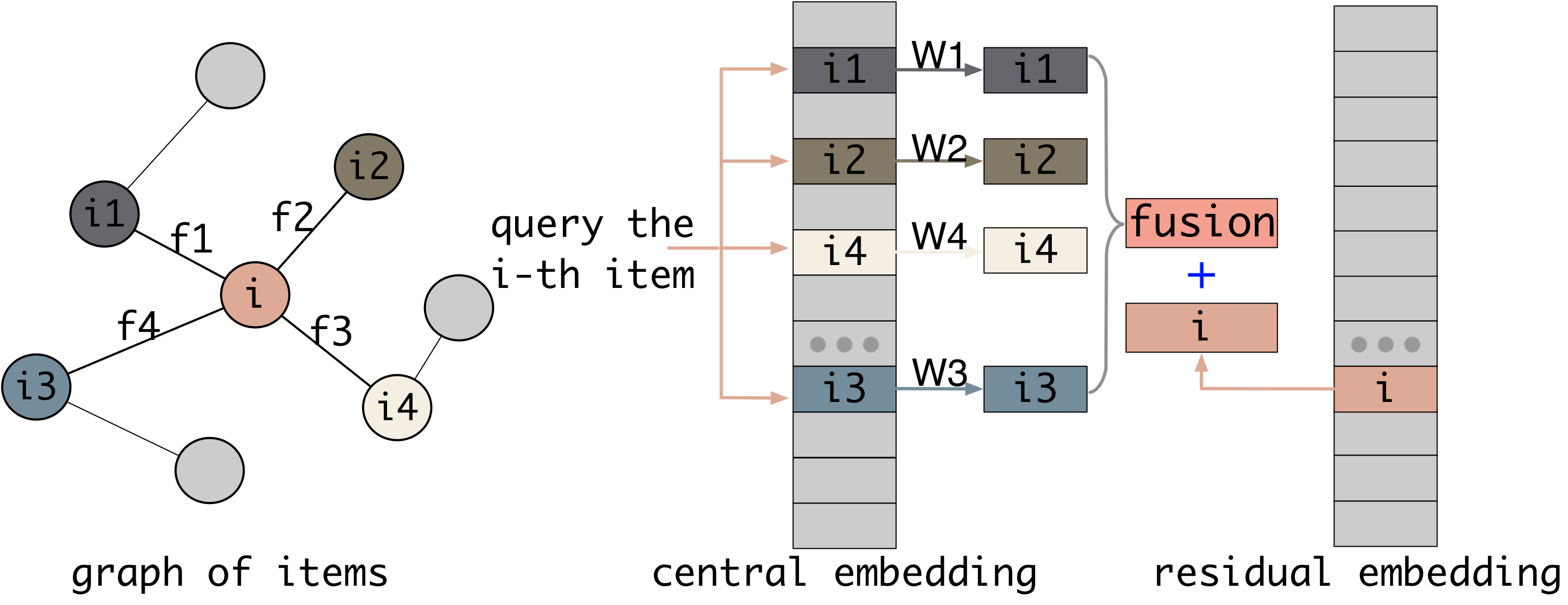}
	%\vspace*{-0.4cm}
	%\includegraphics[scale=0.19]{Graph_and_embedding.png}
	\caption{The left part of this figure is the items graph (Weighted undirected graph) constructed based on users' click behaviors. The weight between the each item in the graph are the co-occurrence frequency in a short-term period of users' click behaviors. When querying the embedding vector of item $i$, we queries the adjacent items $i_1$, $i_2$, $i_3$ and $i_4$ and the connection weights $f_1$, $f_2$, $f_3$ and $f_4$. We combine the central embedding basis vectors of these adjacent in weight $w_1$, $w_2$, $w_3$ and $w_4$ as the final central embedding vector linearly. Weights varies based on different function $g()$. Finally, the central embedding vector is added by residual embedding vector of $i$-th item.}
	\vspace{-0.4cm}
	\label{fig:graph_embedding}
\end{figure}

For instance,  to get the embedding vector of item $i$, the adjacent items list $\{i_1$, $i_2$, $i_3$, $i_4\}$is retrieved from the graph in Figure \ref{fig:graph_embedding}. We select the central embedding basis of items $\{i_1$, $i_2$, $i_3$, $i_4\}$ as the basis of the final central embedding of item $i$. The final central embedding vector of item $i$ is weighted average by the weight $w_1$, $w_2$, $w_3$, $w_4$. The weights $w_1$, $w_2$, $w_3$, $w_4$ are obtained based on the connection weight $f_1$, $f_2$, $f_3$, and $f_4$ in the graph. Finally, the final embedding vector of item $i$ is derived as the sum of its final central embedding vector and residual embedding vector.

We adopt three different forms of function $g$ to fetch the linear combination coefficients matrix $\bm{W}$. The simplest one is the average function which means that all of the linear combination coefficients is the same:
\begin{equation}
\label{avg}
g_{AVG}(\bm{Z}) = {\rm avg_{nz}}(I(\bm{Z} > 0)).
\end{equation}
 The indicator function $I (\bm{Z} > 0)$ means that each element of the $F$ is changed to 1 if it is greater than 0, otherwise changed to 0.
 The ${\rm avg_{nz}}$ operation averages all the non-zero elements of every row vector in the matrix, and the 0 elements remain unchanged.

\begin{comment}

\begin{figure}[h]
	\centering
	\subfigure[Graph connection of the goods, ]{
		\label{fig:graph:a} %% 第一幅图的标签
		\includegraphics[width=0.15\textwidth]{Graph_and_embedding.png}}
	\hspace{1in} \subfigure[Inference process of Res-embedding structure]{
		\label{fig:graph:b} %% 第二幅图的标签
		\includegraphics[width=0.4\textwidth]{graph_embedding_virtue_1.png}}
	\caption{Whole process of the Res-embedding structure. When obtaining the embedding vector of goods $i$.
		The adjacent goods of goods $i$ in the graph is queried first, then it is fused with the central embedding vectors of the adjacent goods. Finally, residual embedding vector is added.}
	\label{fig:graph} %% label for entire figure
\end{figure}
\end{comment}

The second form of function borrows the idea from GCNs. Row normalization and column normalization are performed on adjacent matrix, and the linear combination weight is the connection weight. That is \vspace{-0.2cm} \begin{equation}g_{GCN}(\bm{Z})=\bm{D}^{-\frac{1}{2}}\bm{Z}\bm{D}^{-\frac{1}{2}}.\vspace{-0.1cm}\end{equation}
The degree matrix $\bm{D}\in \mathbb{R}^{N\times N}$ is calculated by row sum of adjacency matrix $\bm{Z}$.

Last form of function is the attention method. We need to introduce $\bm{C_b}$ to calculate the attention score as the linear combination weights, so the $g_{ATT}()$ is rewritten as $g_{ATT}(\bm{Z},\bm{C_b})$

The $g_{ATT}(\bm{Z},\bm{C_b})(i,j)$ means the $i$-th row and $j$-th column element of output of $g_{ATT}(\bm{Z},\bm{C_b})$. $\cdot$ represents the inner product of two vectors. $\bm{C_b}{(i,:)}$ repesents the $i$-th row vector of matrix $\bm{C_b}$.
\begin{equation}
\begin{aligned}
&g_{ATT}(\bm{Z},\bm{C_b})(i,j)= \frac{\exp{(\bm{C_b}{(i,:)}\cdot \bm{C_b}(j,:))}}{\sum_{ k \in \{k|\bm{Z}(i,k)>0\} }{\exp{(\bm{C_b}{(i,:)}\cdot \bm{C_b}(k,:))}}}, \\
&(i,j) \in \{(i,j)|Z(i,j)>0\}
\end{aligned}
\end{equation}

%In order not to affect the inference speed. We use inner product as the score list of attention. Attention scores are used as weighs. 

%The difference between these three different fusion mechanisms is that the embedding vectors of adjacent item is weighted by different weights. 
Beyond averaging, GCNs and attention methods pay more attention on the importance of each adjacent item. Therefore, their performance is supposed to be better, which are reflected in the experiments.

Our optimization objective function is
\begin{equation}\min_{\Theta,\bm{C_b},\bm{R}}E_0(\Theta,\bm{C_b},\bm{R}) + \lambda L(\bm{R}).\end{equation}
$\Theta$ is the parameter set of the deep CTR prediction model except for the embedding layer. $E_0(\Theta,\bm{C_b},\bm{R})$ is the cross-entropy loss of CTR, and the $l_2$ regularization term $L(\bm{R})$ with the coefficient $\lambda$ is added to bound the scale of residual embedding vectors. Proposition 1 in the appendix and related analysis will illustrate that the proposed res-embedding can reduce the distance among the goods belonging to the same interest domain in the embedding space and the goods, while maintaining the distance between the goods in different interest domains.

	\section{Experiments}
\label{Experiments}
In this section, we perform a series of experiments around res-embedding. The generalization experiments verify the theoretical conclusion through training set decrement experiment and visualization experiments. Experiments on Amazon and Movielens datasets illustrate that the proposed embedding layers could enhance the performance of various deep CTR prediction models. 
\vspace{-0.2cm}
\subsection{Datasets and Experimental Setup}
\begin{table}[h]
	
\begin{tabular}{ccccc}
	\hline
	Datasets&Users& Items& Categories&Clicks\\
	\hline
	Amazon-Electronics& 192,403 & 63,001& 801&1,689,188\\
	Amazon-Books&603,669 &367,984 & 1,602&8,898,041\\
	MovieLens&138,493 & 27,278& 21&20,000,263 \\
	\hline

\end{tabular}
	\caption{Statistical information of datasets in this paper. "Clicks" means click number of the dataset, and for Movielens dataset, it means the number of the ratings.}
\end{table}

Amazon Datasets\cite{He2016Ups}: We utilize the product reviews and metadata from Amazon to validate performance of res-embedding. Electronics and Books subsets are used by our experiments. The statistical information of Electronics and Books is shown in Table 2. The complete user behaviors are $(g_1,\dots,g_N)$. Our model is to predict the probability of one item reviewed in the $n+1$-th step under the first $n$ reviewed item. For each user, there is a pair sharing the same historical behavior $(b_1,\dots,b_{n})$ with one positive and one negative sample. The target item of the  positive sample is $b_{n+1}$ and that of the negative sample is randomly sampled from all of the sample. Training and Testing datasets are divided by users.

MovieLens Datasets\cite{Harper2015The}: The statistical information of Electronics and Books is shown in Table 2. We choose the movie that users rate at a certain time as the target movie $m_t$ and $n=50$ movies $(m_1,..., m_n)$ with a score of no less than 3 before this time as the user behavior sequence. Target movie rated more than 3 are labeled as positive, otherwise negative.$\{(m_1,\cdots,m_n),m_t,y\}$ is a CTR prediction sample.

For all models and datasets, we use Adam as the optimizer with exponential decay, in which learning rate is set as 0.1 and batch-size is set as 128. The regularization coefficient $\lambda$ of the residual embedding vectors in res-embedding structure is set as 0.006. We select the windows redius $\Delta =2$ and the $K=8$ mentioned in Algorithm \ref{alg:graph}.

AUC\cite{Fawcett2005An} is adopted as criteria which measures the goodness of order by ranking all the samples with predicted CTR. We utilize res-embedding on several different deep models to validate the performance of res-embedding. These deep models are listed as follows:

Basic MLP: The basic MLP is the original MLP. It receive the embedding vector of target item and the sum of embedding vectors of the items clicked by users historically as the input. In this experiments, layers of MLP are set as 36$\times$400$\times$120$\times$2.

PNN\cite{Qu2017Product}: PNN is an variant of the basic MLP, and the difference between them is that the PNN receives the product the sum embedding vectors and target product as the an extra input vector. The layers of PNN are set as 54$\times$400$\times$120$\times$2.

DIN\cite{Cheng2016Wide}: DIN introduces the attention mechanism into the CTR model. It extracts several interest vectors of the users from their historical behaviors. The embedding vector of the target item is extracted into an interest vector and adopt the attention operation with the interest vector of the behavior sequence, then input the result to the basic MLP.
\begin{comment}
An variation of user weighted AUC which measures the goodness of intra-user order by averaging AUC over users and is shown to be more relevant to online performance in display advertising system. We adapt this criteria in our experiments. it is calculated as follows:
$$GAUC=\frac{\sum_i{AUC_i*N_i}}{\sum_i{N_i}}$$

In order to indicate the relative difference of AUC over different model, RelaImpr metric follow \cite{yan2014coupled}  is defined as below:
$$Rel aImpr = (\frac{AUC(measured model)-0.5}{AUC(base model)-0.5}-1)\times 100\%$$
\end{comment}

\subsection{Validation of Theorem 1}

\begin{figure}[h]
	\setlength{\abovecaptionskip}{0.02cm}
	\centering
	\includegraphics[scale=0.5]{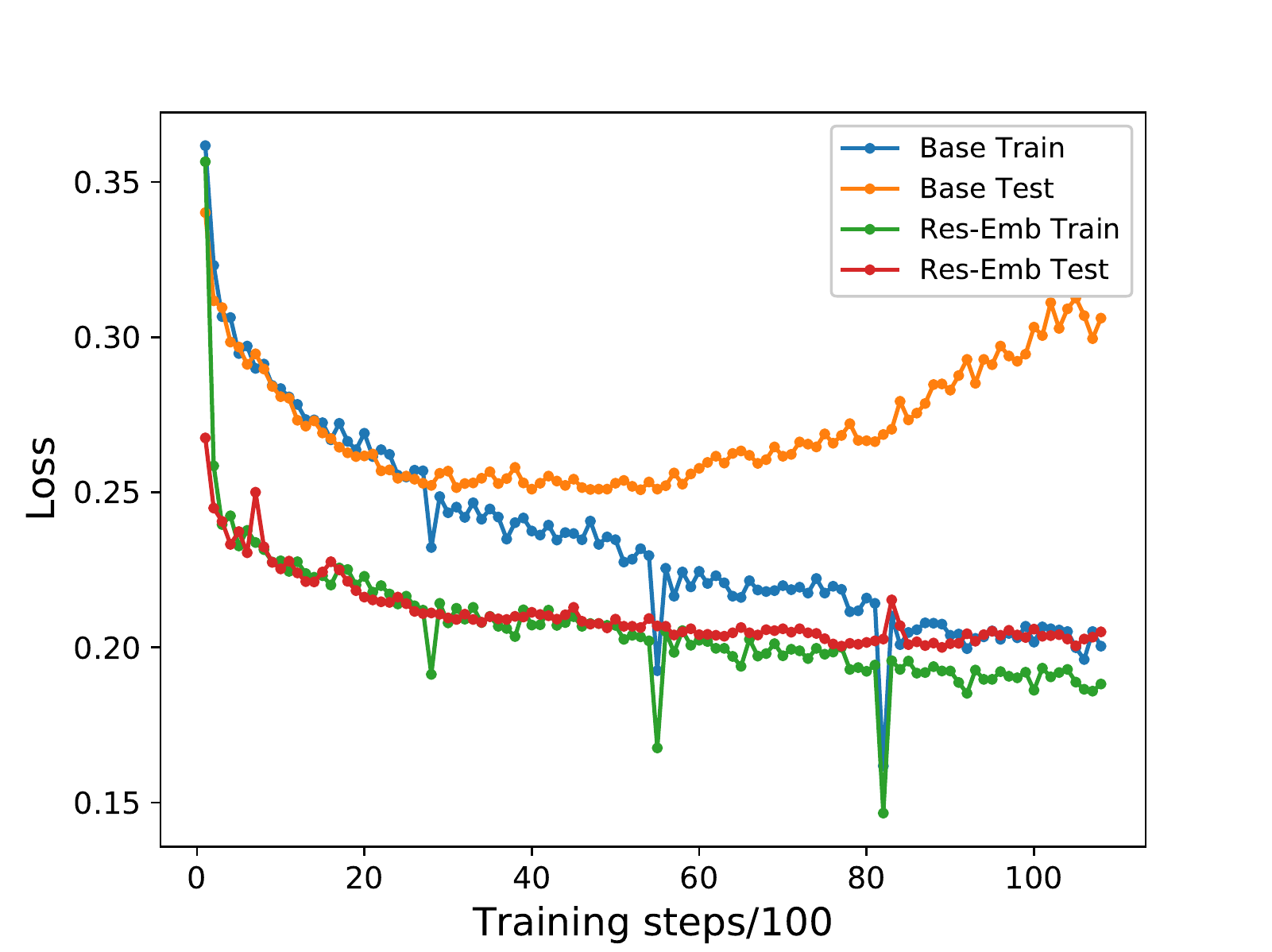}
	\caption{Compare the original embedding and the res-embedding with attention in DIN model for electronics dataset, The horizontal axis is the number of training steps/100, and the longitudinal axis is the loss function of the CTR model. The lines of different colors represent the loss function curves of different models under different data. For example, "Base Train" represents the line of the original embedding structure under the training dataset.}
	\vspace{0.2cm}
	\label{fig:train_test}
\end{figure}

We compare the train-test line of the original embedding layer and that of the res-embedding with attention in DIN model for electronics dataset. As shown in Figure \ref{fig:train_test}, under the original embedding structure, the CTR task has a very strong over fitting phenomenon. That is, as the number of training steps increases, the gap between training and test loss group quite quickly. When the res-embedding is adopt as the input of model, this error has been greatly reduced, which indicates that the generalization error bound of CTR prediction model is effectively controled by res-embedding structure.
\begin{figure}[t]
	\setlength{\abovecaptionskip}{-0.02cm}
	\centering
	\includegraphics[scale=0.5]{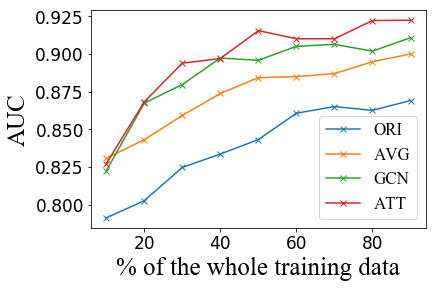}
	\caption{The proportion of to whole training set varies from 10\% to 90\%. ORI, AVG, GCN, ATT represent the original embedding, res-embedding with average, GCN, and attention fusion mechanism.}
	\vspace{-0.2cm}
	\label{fig:decay}
\end{figure}

\begin{figure}[!ht]
	\setlength{\abovecaptionskip}{-0.02cm}
	\centering 	
	\subfigure[ORI]{\label{fig:vitual_ori}
		\includegraphics[width=0.4\linewidth]{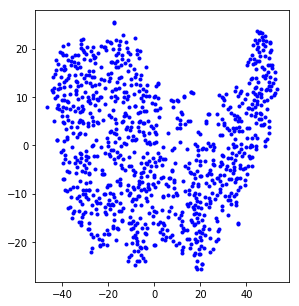}}	
	\hspace{0.01\linewidth}	
	\subfigure[AVG]{\label{fig:vitual_avg}	
		\includegraphics[width=0.4\linewidth]{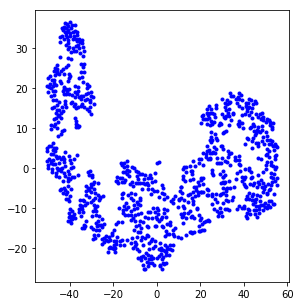}}
	\vfill
	\subfigure[ATT]{\label{fig:vitual_att}
		\includegraphics[width=0.4\linewidth]{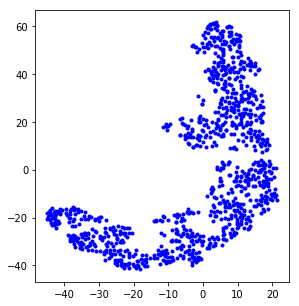}}
	\hspace{0.01\linewidth}
	\subfigure[GCN]{\label{fig:vitual_gcn}
		\includegraphics[width=0.4\linewidth]{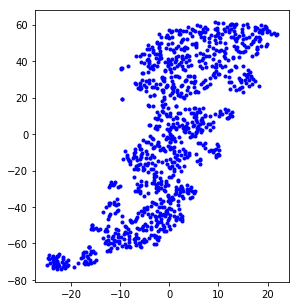}}
	\caption{The blue dots denote embedding vectors of the top 1000 high frequency items. ORI represents the original embedding structure, AVG, ATT and GCN are res-embedding with three different fusion mechanisms. }
	\vspace{-0.4cm}
	\label{fig:vitual}
\end{figure}
%10\% to 90\% 

Aiming to verify the generalization performance promotion of res-embedding further, we compared original embedding structure with res-embedding in DIN model under the decay datasets. We reduce scale of the training set of Books dataset and utilize them to train the deep CTR model and the proportion of training dataset to whole training dataset decays from 90\% to 10\%. In Figure \ref{fig:decay}, we find that only adapting 20\% or 30\% dataset to train with res-embedding could obtain the similar performance to using the whole training set. This phenomenon demonstrates that res-embedding structure can partly relieve the problem of generalization performance loss due to insufficient training data.

Theorem \ref{Theorem2} tells us that the group aggregating embedding vectors are helpful to improve generalization performance. In order to prove the consistency between res-embedding and theoretical analysis, we visualize embedding vectors of the top 1000 highest frequency items in the deep CTR model with t-sne visualization methods. Figure \ref{fig:vitual_ori} shows visualized embedding vectors of original embedding structure. It can be seen that embedding vectors are randomly located in the whole region. While in Figure \ref{fig:vitual_avg}, \ref{fig:vitual_gcn}, and \ref{fig:vitual_att}, the density of embedding distribution is not uniform, which is different from the original embedding structure.  Embedding vectors in some parts are dense and in other parts are sparse. In another word, embedding vectors are more aggregated locally, which is consistent with our theory.\vspace{-0.2cm}

\subsection{Total Promotion of Res-embedding}
 \begin{table}[h]
 	\label{table_result}
	\centering
	\small{
	\begin{tabular}{ccccc}
		\toprule  %添加表格头部粗线
		Methods & AUC(Electronics)& AUC(Books)& AUC(Movielens) \\
		\midrule  %添加表格中横线
		MLP& 0.8558$\pm$0.0004& 0.8833$\pm$0.0013 &0.7242$\pm$0.0002\\
		MLP+R& 0.8649$\pm$0.0008& 0.8802$\pm$0.0021&0.7337$\pm$0.0003\\
		MLP+Skip&0.8710$\pm$0.0006&0.8964$\pm$0.0013&0.7275$\pm$0.0003\\
		MLP+n2v&0.8756$\pm$0.0005&0.8973$\pm$0.0010&0.7292$\pm$0.0002\\
		MLP+sms&0.8695$\pm$0.0008&0.9172$\pm$0.0002&0.7270$\pm$0.0003\\
		MLP+AVG& 0.8857$\pm$0.0008& 0.8977$\pm$0.0018&0.7326$\pm$0.0001\\
		MLP+GCN&\textbf{0.8917$\pm$0.0006}&0.9240$\pm$0.0028 &\textbf{0.7389$\pm$0.0003}\\
		MLP+ATT& 0.8907$\pm$0.0011& \textbf{0.9295$\pm$0.0000}&0.7385$\pm$0.0004\\
		\midrule  %添加表格中横线
		PNN& 0.8601$\pm$0.0005& 0.8930$\pm$0.0017&0.7311$\pm$0.0004\\
		PNN+R& 0.8671$\pm$0.0002& 0.8932$\pm$0.0009&0.7363$\pm$0.0002\\
		PNN+Skip&0.8771$\pm$0.0006&0.9006$\pm$0.0012&0.7313$\pm$0.0004\\
		PNN+n2v&0.8817$\pm$0.0004&0.9041$\pm$0.0013&0.7327$\pm$0.0004\\
		PNN+sms&0.8758$\pm$0.0002&0.9182$\pm$0.0015&0.7330$\pm$0.0005\\
		PNN+AVG& 0.8930$\pm$0.0003&0.9052$\pm$0.0009&0.7449$\pm$0.0004\\
		PNN+GCN& 0.9042$\pm$0.0006&0.9325$\pm$0.0023 &\textbf{0.7458$\pm$0.0003}\\
		PNN+ATT& \textbf{0.9057$\pm$0.0003}&\textbf{ 0.9357$\pm$0.0011}&0.7449$\pm$0.0006\\
		\midrule  %添加表格中横线
		DIN& 0.8635$\pm$0.0000&0.8971$\pm$0.0003&0.7288$\pm$0.0001\\
		DIN+R& 0.8752$\pm$0.0000&0.8902$\pm$0.0002&0.7360$\pm$0.0001\\
		DIN+Skip&0.8786$\pm$0.0001&0.9048$\pm$0.0001&0.7303$\pm$0.0000\\
		DIN+n2v&0.8832$\pm$0.0001&0.9078$\pm$0.0001&0.7319$\pm$0.0000\\
		DIN+sms&0.8802$\pm$0.0002&0.9285$\pm$0.0001&0.7275$\pm$0.0001\\
		DIN+AVG&0.8981$\pm$0.0003& 0.9052$\pm$0.0003&0.7383$\pm$0.0002\\
		DIN+GCN&0.9090$\pm$0.0003&0.9372$\pm$0.0001  &\textbf{0.7429$\pm$0.0001}\\
		DIN+ATT&  \textbf{0.9106$\pm$0.0003}&\textbf{0.9404$\pm$0.0003} &0.7412$\pm$0.0000\\
		\bottomrule %添加表格底部粗线
	\end{tabular}}
	\caption{The AUC results on amazon datasets (Electronics and Books) and movielens dataset. model without any addition represents the deep model with original embedding layer. R represents the deep model with original embedding layer with regularization constraints. The AVG, GCN, ATT represent the fusion mechanism under res-embedding structure: average, GCN, attention. The MLP+Skip\cite{barkan2016item2vec},n2v\cite{Grover2016node2vec} and sms\cite{Parsana2018Improving} are 3 pretrain methods, means Skip-gram, node2vec, and siamese auxiliary network.}
	\vspace{-0.2cm}
\end{table} 

Table 3 shows the experimental results on two Amazon Datasets and MovieLens Datasets. All the results contain the mean AUC and its variance under experiments repeated for 5 times. Comparing with the original embedding structure with L2-regularization term and without it, we find that the performance has been improved after adding regularization terms, but the promotion is not so large. As mentioned in the theoretical analysis, it is because reducing the scale of embedding as a whole limits the expressive power of embedding while reducing $R_{max}$. Res-embedding structure promotes the AUC of the deep model under the CTR task. Compared with some other embedding methods like Skip-gram, node2vec and siamese auxiliary network, res-embedding also acheive better performance. The average function promotes less than the attention and GCN function. The main reason is that the average function only aggregates the nearest neighbors indiscriminately while the other two methods take account of the different characteristics of different neighboring items.\vspace{-0.2cm}

 \subsection{Comparison of parameter quantities}
  \begin{table}[ht]
 		\label{dimension}
 	\centering
 	\small{
 		\begin{tabular}{ccc}
 		
 			\toprule  %添加表格头部粗线
 			
 			Models      & Dim of embedding  & AUC(Electronics) \\
 			\midrule  %添加表格中横线
 			DIN  & 18  & 0.8635     \\
 			DIN  & 36  & 0.8751     \\
 			\midrule  %添加表格中横线
 			DIN+AVG& 18 & 0.8981     \\ 
 			DIN+AVG& 36 & 0.9105     \\ 
 			\midrule  %添加表格中横线
 			DIN+GCN& 18 & 0.9090     \\ 
 			DIN+GCN& 36 & 0.9115     \\ 
 			\midrule  %添加表格中横线
 			DIN+ATT& 18 & 0.9106     \\ 
 			DIN+ATT& 36 & 0.9129    \\
 			\bottomrule 
 	\end{tabular}}
 	
 	\caption{For the electronic dataset, the AUC of the DIN model uses different embedding methods under different dimensions of embedding vector. Dim of embedding means the dimension of the embedding vectors.}
 	\vspace{-0.2cm}
 \end{table}

In Table 3, we ensure the embedding space of comparative method and res-embedding are the same. However, res-embedding uses twice parameters compared with the original embedding structure, so we are supposed to discuss the effect of changing of the parameters and embedding dimensions on performance. As shown in Table 4, with the increase of embedding dimension(compared with DIN models with 18 dim and DIN model with 36dim), the performance of DIN is improved, which indicates that increasing the dimension of embedding will at least not loss performance. Focusing on DIN models with 36dim, DIN+AVG model, DIN+GCN model and DIN+ATT model with 18dim, which means that res-embedding has the same size of parameters with original embedding structure and has smaller dimensions, our method is still greatly improved. In summary, although res-embedding uses twice as many parameters as baseline, the improvement of res-embedding is due to the improvement of the method itself, not the increase of the parameters. Even if the parameters of baseline are increased to the same as res-embedding, and the embedding dimension of baseline is also higher than res-embedding, res-embedding still has a significant improvement.

 \subsection{Effectiveness of residual part}
  \begin{figure}[h]
   \setlength{\abovecaptionskip}{0.0cm}
 	\centering
 	\includegraphics[scale=0.5]{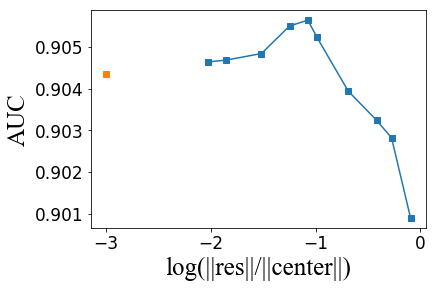}
 	\caption{The relationship between the scale of the residual part and the AUC based on the electronics dataset. The orange dot means the situation that the residual embedding disappears. The deep model is DIN with res-embedding in attention.}
 	\label{fig:pathdemo}
 \end{figure}
 In this subsection, we will verify the effectiveness of residual embedding vectors. The experimental results are shown in Figure \ref{fig:pathdemo}, which is based on the electronics dataset, res-embedding in attention technology with DIN. In experiments, we increase the scale of residual embedding by controlling the regularization coefficient of residual embedding matrix. As the scale ratio between residual and central embedding vector varies from large to small, the performance is gradually improved. When the scale ratio between residual and central embedding is reduced apoach to 1:10, the AUC criteria reaches its peak value. The promotion of performance is consistent with our theory for the smaller scale of residual embedding means smaller envelope radius of items with the same interest domain in the embedding space, which will improve the generalization performance. However, while continuing to reduce scale of the residual to disappear, the AUC criteria drops slightly, which may be attributed that over-small scale of the residual embedding vector reduces the representation ability of the embedding layer. In summary, the residual part can be considered as an adjuster to balance generalization and representation ability.

	\section{Conclusion}
In this paper we propose a novel res-embedding structure to improve the generalization of deep CTR prediction task. We theoretically proved that the generalization error of the depth CTR model is related to the aggregation of embedding packets. We use the user historical behaviors to construct the graph of items and calculate the central embedding based on this graph. Several fusion methods are adopt to generate central embedding. Residual embedding is used to control the degree of embedding aggregation. In the experiment, we validate the effectiveness of the residual structure on the CTR task public datasets. We also observed a significant improvement in generalization performance, and the visualization experiments also verified the changes in embedding structure. The graph among items is pre-generated so far, which may be affected by changes of sample distribution. In the future, we will consider using an end-to-end approach to dynamically update the graph.
	\bibliographystyle{ACM-Reference-Format}
	\bibliography{res-embedding}
	
\end{document}

\bibliographystyle{plain}
Firstly, we give the definition of robustness, and give a series of lemmas including the robustness of MLP, the concentration inequality of polynomial distributions and so on. Then we prove Theorem \ref{Theorem1}, that is, derive the generalization bounds based on single-clicked historical behavior. Next, we prove Theorem \ref{Theorem2}, which is to derive generalization bounds based on the multi-clicked historical behavior sequence. Finally, we prove the Proposition \ref{propostion 1}, which implies the consistency between the method and theory.

We first define some symbols. The number of interest implicit state $z$ is $N_z$. The envelope radius of set $V\subset \mathbb{R}^n$ is $R_0\in \mathbb{R}^{+}$. 
That is, $\exists s\in \mathbb{R}^n$ and $\forall w \in V$ 
, there will be 
$\|w-s\|_2 < R_0$ defined as $V\in \phi(R_0)$. $r$-covering set $K$ means $\forall x,y\in K$, the $\|x-y\|_2\leq r$. $r$-covering disjoint sets $K_1$ and $K_2$ mean that they are both $r$-covering sets and $K_1\bigcap K_2=\emptyset$ .

\section{Preparation}
\begin{Definition}
	(Robustness)  For a learning algorithm $\mathcal{A}$, and sample $s=\{x,y\}$. $x\in \mathcal{X} $ and $y\in \mathcal{Y}$. $l(\mathcal{A}, s)$ represents the loss function $l(f_{\mathcal{A}}(x),y)$ of model $f_{\mathcal{A}}():\mathcal{X}\rightarrow\mathcal{Y}$ optimized by algorithm $\mathcal{A}$. The domain of the sample $s$ is $\{\mathcal{X},\mathcal{Y}\}$. It can be partitioned into $L\in \mathbb{N}$ disjoint sets $K_1,K_2,\cdots ,K_{L}$, and $\forall s_1,s_2 \in K_i$,$\forall i =1,\cdots,L$ , there satisfies $|l(\mathcal{A},s_1)-l(\mathcal{A},s_2)|\leq \epsilon$, then the algorithm $\mathcal{A}$ will be defined as $(L,\epsilon)$-robust.
\end{Definition}
This definition of robustness is from \cite{Xu2012Robustness}. The robustness bound $\epsilon$ and the number of the disjoint sets $L$ decided the robustness of the algorithm. Intuitively, the large $L$ or large $\epsilon$ robust model will cause the fact that output is greatly influenced by input, which will cause the output gap between training set and testing set. Previous literature proves that the generalization bound of the learning algorithm consists of  $\epsilon$ and $L$.

\begin{Lemma}
	\label{lemma}
	$\forall$ $\Gamma \subset \mathbb{R}^d$ with the radius $R_0\in \mathbb{R}$. That is, $\Gamma \in \phi(R_0)$. There are $L = (2R_0\sqrt{d}/r)^d$ $r$-covering disjoint sets of $\phi(R)$, $K_1,\cdots,K_L \subset \phi(R)$ and $\bigcup_{i=1}^{L} K_i = \phi(R_0)$. $\forall x,y\in K_i$, the $\|x-y\|_2\leq r$.
\end{Lemma}
The Lemma \ref{lemma} is a conclusion derived from the Definition 27.1 and Example 27.1 of \cite{Shalev2014Understanding}. It points out the number of subsets that can be divided into $r$ diameter for a region with a radius of $R_0$.

\begin{Lemma}
	\label{lemma2}
	The $s=\{x_1,\cdots,x_n,y^*\}$ is training sample of $D$ layers MLP with ReLU nonlinear function with concatenation of $\{x_1,\cdots,x_n\}$ and $x_i \in \mathcal{X}_i \subset \mathbb{R}^d$ for $\forall i$ as input, $y^*\in \mathcal{Y}=\{0,1\}$ as the label. The loss function of MLP is $|f(x_1,x_2,\cdots,x_n)-y^*|$. Its forward process is
	\begin{equation}
\begin{aligned}
h^0&=[x_1,\cdots,x_n],h^t = ReLU(W_th^{t-1}+b_t),\\
y &= sigmoid(W_Dh^{D-1}+b_{D}).
\end{aligned}
\end{equation}
	$h^0\in \mathbb{R}^{nd}$ and $y\in \mathbb{R}$.  If $\mathcal{X}_i \in \phi(R_i)$, and the parameter sets $\{W_t,b_t\}$ for $\forall t$ have been trained by the alogrithm $\mathcal{A}$, the alogrithm is $\mathcal{A}$ is $(\|W\|_2^Dr\sqrt{n}$,$2(2R_{max}\sqrt{d}/r)^{nd})$-robust.
	 Among them, $R_{max}$ is the maximal value in ${R_1,\cdots,R_n}$, and 
	$\|W\|_2= \sum_{t=1}^{D}\|W_t\|_2/D$.
	In addition, any $s = \{x, y\}, s'= \{x',y'\}$, as long as $y=y'$. $|l(f,s)-l(f,s')|\leq\|W\|_2^D\|x'-x\|_2$ is satisfied.

\end{Lemma}

\begin{Proof}
	For each $\mathcal{X}_i\in \phi(R_i)$, there are $L_i=(2R_i\sqrt{d}/r)^d$ $r$-covering disjoint sets $K_1^i,\cdots,K_{L_i}^i \subset \mathcal{X}_i $ according to Lemma \ref{lemma}. 
	
	If two samples $s_1 = \{x^1,y^1\}$ and $s_2 = \{x^2,y^2\}$ satisfy that
	$y^1=y^2=y$ and $\forall i=1,\cdots,n$, $ x_i^1$, $x_i^2 \in K_{l_i}^i$, we define that $s_1,$ $s_2\in \mathcal{K}(l_1,\cdots,l_n,y)$. There exist $N =2\prod_{i=1}^n L_i= 2\prod_{i=1}^n(2R_i\sqrt{d}/r)^d=2(2R_{max}\sqrt{d}/r)^{nd}$ disjoint sets $\mathcal{K}()$
	
	 $\forall s_1,$ $s_2\in \mathcal{K}(l_1,\cdots,l_n,y)$. $\forall i$, $K_{l_i}^i $ is $r$-covering set. Therefore, the 
	 		\begin{equation}
	 		\label{distance }
	 \|x_i^1-x_i^2\|_2 \leq r.
	 \end{equation}
	 
The difference between their loss functions is 
		\begin{equation}
	\begin{aligned}
	\label{first}
	&|l(f,s_1)-l(f,s_2)|=|y^1-f(x^1)|-|y^2-f(x^2)|\leq|f(x^1)-f(x^2)|| \\
	&= |a_D(W_Dh^D_1+b_D)-a_D(W_Dh^D_2+b_D)|\leq \beta_D\|W_D\|_2\|h^D_2-h^D_1\|
	\end{aligned}
	\end{equation}
	The $\beta_t=max\frac{|a_t(x)-a_t(x')|}{|x-x'|}$ for $\forall x, x'\in$ domain$a_t$ , and $a_t()$ mean activation function on $t$-th layer.
	For the $t$-th layer 
		\begin{equation}
	\begin{aligned}
	\label{ditui}
	&\|h^t_2-h^t_2\|_2\leq \|a_D(W^th^{t-1}_1+b_t)-a_D(W^th^{t-1}_2+b_t)\|_2 \\
	&\leq \beta_t\|(W^th^{t-1}_2-W^th^{t-1}_2)\|_2 \leq \beta\|W^t\|_2\|h^{t-1}_1-h^{t-1}_2\|_2.
	\end{aligned}
	\end{equation}
	The equation(\ref{ditui}) is the recurrence relation between the 2-norm of the hidden state difference of layer $t$ and the 2-norm of the $t-1$ level hidden state.
	According to equation (\ref{distance }), there will be 
	\begin{equation}
	\label{final_distance}
	\|x^1-x^2\|_2=\sqrt{\sum_{i=1}\|x^1_i-x^2_i\|^2_2}\leq r\sqrt{n}.
	\end{equation}
	Therefore, (\ref{first}) combination (\ref{ditui}) and (\ref{final_distance})the difference of the loss function is 
			\begin{equation}
	\begin{aligned}
	\label{almost_final}
	&|l(f,s_1)-l(f,s_2)|\leq \beta_D\|W_D\|_2\|h^D_2-h^D_1\|_2\leq\\
	& \prod_{i}\beta_i\|W_i\|_2\|h^0_2-h^0_1\|_2\leq \prod_{i}\beta_i\|W_i\|_2\|h^0_2-h^0_1\|_2\leq\\ &\prod_{i}\|W_i\|_2\|x_2-x_1\|_2\leq r\sqrt{n}\prod_{i}\beta_i\|W_i\|_2.
		\end{aligned}
	\end{equation}
	If $a_t()=ReLU()$, then $\beta_t = 1$ and if $a_t() = sigmoid()$, then $\beta_t < 1$. Combining the mean inequality $(\prod_{i}\beta_i\|W_i\|_2)^{1/D}\leq \frac{\sum_{i}\|W_i\|_2}{D}$ with (\ref{almost_final})
	\begin{equation}
	\label{final_lemma}
	|l(f,s_1)-l(f,s_2)|\leq(\sum_{i}^{n}\frac{\|W_i\|_2}{D})^D\|x^1-x^2\|_2\leq r\sqrt{n}(\sum_{i}^{n}\frac{\|W_i\|_2}{D})^D
	\end{equation}
	 According to the definition of the robustness. The $D$-th with ReLU activation function is $(\|W\|_2^Dr\sqrt{n}$
	 ,$2(2R_{max}\sqrt{d}/r)^{nd})$-robust. The first inequality in (\ref{final_lemma}) also obtained the additional conclusion of Lemma \ref{lemma2}.
		
\end{Proof}

\begin{Lemma}
	\label{lemma3}
	(Breteganolle-Huber-Carol inequality)The random variable $x$ belongs to domain$X$ with $K$ disjoint sets $C_1,\cdots,C_K$. $ \bigcup C_i = $domain$X$, let $n_i$ be the number of points that fall into the region $C_i$, $n$ is the total number. there will be
	$$
	P(\sum_{i=1}^{K}|\frac{n_i}{n}-P(x\in C_i)| \geq \lambda)\leq 2^K\exp(\frac{-n\lambda^2}{2})
	$$
\end{Lemma}

This is the concentration inequality of multinomial distribution from Proposition A6.6
of \cite{Vaart1996Weak}

\section{Generalization bounds based on single-clicked historical behavior}
\begin{Assumption}
	\label{Assumption1}
	A training sample $s = \{x_h,x_t,y^*\}$ consists of clicked items $x_h$, target item $x_t$ and label $y^*$. The interest hidden states of clicked items and target item $z_h$ and $z_t$ are sampled from two distribution $P_o(z)$ and $P_t(z)$ respectively. The clicked items $x_h$ and target item $x_t$ are sampled from the conditional distribution $P(x|z)$. The domain of $P(x|z=i)$ is abbreviated as domain$P_i$. The label $y^*$ of data is sampled from the set $\{0,1\}$.
\end{Assumption}

\begin{Theorem}
	\label{Theorem1}
	Sampling $N$ training data under Assumption \ref{Assumption1}, $s=\{x_h,x_t,y^*\}$. For the $D$ layers MLP with ReLU $y=f(x_h,x_t)$ and the loss function $l(f,s)=|f(x_h,x_t)-y^*|$ defined the same as Lemma \ref{lemma2}. With the probability $1-\delta$, there will be\vspace{-0.3cm}
	\begin{equation}
	\label{equ1}
	\begin{aligned}
	&|E_s(l(f,s))-\frac{1}{N}\sum_{i=1}^{N}l(f,s_i)|\leq \\
	&\inf_r{\sqrt{2}\|W\|_2^Dr+l_M\sqrt{\frac{4N_z^2(R_{max}\sqrt{d}/r)^{2d}\ln{2}+2ln(1/\delta)}{N}}}.
	\end{aligned}
	\end{equation}
	Among them, $l_M$ is the maximum value of $l(f,s)$. $d$ is dimension of each input vector, domain $P_i\in \phi(R_{max})$ for all $z=i$. $\|W\|_2$ is the average of 2-norm of all parameter matrices.
\end{Theorem}

\begin{Proof}
	\label{Proof1}
	According to Assumption \ref{Assumption1}, each sample $s= \{x_h,x_t,y\}$ exists a behavior interest hidden state $z_h$ and target interest hidden state $z_t$. 
	
	According to the Assumption \ref{Assumption1}, when $z_h=i$ and $z_t=j$ are determined, the domain of $x_h$ and $x_t$ is identified as domain$P_i\in \phi(R_i)$  and domain$P_j\in \phi(R_i)$. According to Lemma \ref{lemma}, domain$P_i$ and domain$P_i$ exist $L_i = (2R_{i}\sqrt{d}/r)^d$ and $L_j = (2R_{j}\sqrt{d}/r)^d$ r-covering disjoint sets $\{K_{1}^i,\cdots,K_{L_i}^i\}$ and $\{K_{1}^j,\cdots,K_{L_j}^j\}$.

	For one sample $s = \{x_h,x_t,y^*\}$ from $N$ training samples, we define the situation that $z_h = i$, $z_t= j$, $x_h \in K^i_{\alpha}$, $x_t \in K^j_{\beta}$, and $y^* = q \in\{0,1\}$ as $s \in \Psi = \psi (i,j,\alpha,\beta,q)$. When $i$ and $j$ remain unchanged, there are $L_i\times L_j \times 2$ cases in this set. Therefore, the total number of the set $\psi (i,j,\alpha,\beta,q)$ is 
	\begin{equation}
	\label{N}
	N_{\psi } = \sum_{i=1}^{N_z}\sum_{j=1}^{N_z} 2L_iL_j\leq\sum_{i=1}^{N_z}\sum_{j=1}^{N_z}2(2R_{max}\sqrt{d}/r)^{2d}  \leq 2N_z^2(2R_{max}\sqrt{d}/r)^{2d}
	\end{equation}
	The difference between expected loss and empirical loss is
	\begin{equation}
	\begin{aligned}
	\label{main}
	&|E_s(l(f,s))-\frac{1}{N}\sum_{i=1}^{N}l(f,s_i)|\leq \\
	&|E_s(l(f,s))-\frac{1}{N} \sum_{\Psi}n_{\Psi}E_{s_{\Psi}}(l(f,s_{\Psi})|s_{\Psi}\in \Psi)| +\\
	&|\frac{1}{N} \sum_{\Psi}n_{\Psi}E_{s_{\Psi}}(l(f,s_{\Psi})|s_{\Psi}\in \Psi) -\frac{1}{N}\sum_{i=1}^{N}l(f,s_i)| 
	\end{aligned}
	\end{equation}

	In equation (\ref{main}) we add and subtract one item $\frac{1}{N} \sum_{\Psi}n_{\Psi}E_{s_{\Psi}}(l(f,s_{\Psi})|s_{\Psi}\in \Psi)$. Among them, $n_{\Psi}$ is the number of samples belonging to $\Phi$ in $N$ training samples. $E_{s_{\Psi}}(l(f,s_{\Psi})|s_{\Psi}\in \Psi)$ is the expectation of the loss function when the sample belongs to $\Psi$.
	
	Based on the definition of expectation, $E_s(l(f,s)) = \sum_{\Psi}E_{s_{\Psi}}(l(f,s_{\Psi})|s_{\Psi} \in \Psi )P(s_{\Psi}\in \Psi)$. $\frac{1}{N}\sum_{i=1}^{N}l(f,s_i) = \frac{1}{N}\sum_{\Psi}\sum_{s_i \in \Psi} l(f,s_i)$. We put them in equation (\ref{main}).
	\begin{equation}
	\begin{aligned}
	\label{main2}
	&|E_s(l(f,s))-\frac{1}{N}\sum_{i=1}^{N}l(f,s_i)| \leq \\
	&\sum_{\Psi}|E_{s_{\Psi}}(l(f,s_i)|s_{\Psi} \in \Psi )||P(s_{\Psi}\in \Psi) - \frac{n_{\Psi}}{N}|+\\
	&\frac{1}{N}\sum_{\Psi}|n_{\Psi}E_{s_{\Psi}}(l(f,s_{\Psi})|s_{\Psi} \in \Psi )-\sum_{s_i \in \Psi}l(f,s_i)|\leq \\
	&l_M\sum_{\Psi}|P(s_i\in \Psi) - \frac{n_{\Psi}}{N}| + \frac{1}{N}\sum_{\Psi}\sum_{s_i \in \Psi}|E_{s_{\Psi}}(l(f,s_{\Psi})|s_{\Psi} \in \Psi)-l(f,s_i)|
	\end{aligned}
	\end{equation}
	$l_M$ is the maximal value of $l(f,s)$.
	According to the concentration inequalities for multinomial distribution in Lemma \ref{lemma3}, in this situation, $K = N_{\Psi}$. Therefore,
	$P(\sum_{\Psi}|P(s_i\in \Psi) - \frac{n_{\Psi}}{N}| \leq \lambda) \geq 1 - 2^{N_{\Psi}}\exp(\frac{-n\lambda^2}{2})$. Let $\delta = 2^{N_{\Psi}}\exp(\frac{-n\lambda^2}{2})$, then $\lambda = \sqrt{\frac{2N_{\Psi}ln2+2ln(1/\delta)}{2}}$. It means that 
	\begin{equation}
	\label{concentrate}
	\sum_{\Psi}|P(s_i\in \Psi) - \frac{n_{\Psi}}{N}| \leq \sqrt{\frac{2N_{\Psi}ln2+2ln(1/\delta)}{n}}
	\end{equation}
	with at least probability $1-\delta$.

	Lemma 2 and its proof show that if $D$-layer MLP input are bound in a region of $\|x-x'\|_2$, then the difference of loss function will not exceed $\|W\|_2^D\|x-x'\|_2$. 
	
	$\forall a,s \in {\Psi}$, $a =\{x_h^1,x_t^1,y^1\}$,$s =\{x_h^2,x_t^2,y^2\}$. $x_h^1$ and $x_h^2$ belong to the same $r$-covering disjoint set. $\|x_h^1-x_h^2\|\leq r$ ,$\|x_t^1-x_t^2\|\leq r$, and $y^1=y^2$. There will be $|l(f,a)-l(f,s)|\leq\|W\|_2^D\sqrt{\|x^1_h-x^2_h\|^2+\|x^1_t-x^2_t\|^2}\leq \sqrt{2}\|W\|_2^Dr$, 
	\begin{equation}
	\label{robust}
	|E_{s_{\Psi}}(l(f,s_{\Psi})|s_{\Psi} \in \Psi)- l(f,s_i)|\leq \sqrt{2}\|W\|_2^Dr.
	\end{equation}
	We substitute (\ref{concentrate}) and  (\ref{robust})   in (\ref{main2}).
	\begin{equation}
	\begin{aligned}
	|E_s(l(f,s))-\frac{1}{N}\sum_{i=1}^{N}l(f,s_i)| \leq \sqrt{\frac{2N_{\Psi}ln2+2ln(1/\delta)}{n}} + \sqrt{2}\|W\|_2^Dr
	\end{aligned}
	\end{equation}
	$r$-covering disjoint sets is a method of partitioning, so $r$ can be adjusted arbitrarily. Therefore, the generalization error bound should be smaller than infimum under all $r$. According to (\ref{N}) and (\ref{concentrate}), with at least probability $1-\delta$.
	\begin{equation}
	\begin{aligned}
	&|E_s(l(f,s))-\frac{1}{N}\sum_{i=1}^{N}l(f,s_i)| \leq \\
	&\inf_r\{\sqrt{\frac{4N_z^2(2R_{max}\sqrt{d}/r)^{2d}ln2+2ln(1/\delta)}{N}} + \sqrt{2}\|W\|_2^Dr\}
	\end{aligned}
	\end{equation}
	
\end{Proof}

\section{Generalization bounds based on multi-clicked historical behavior}
\begin{Assumption}
	\label{Assum2}
	For simplicity, we assume all $p$ periods are $T$ in length. Sample $s = (\{x^1,\cdots,x^{T\times p}\}_h,x_{t},y^*)$ consists of clicked items $\{x^1,\cdots,x^{T\times p}\}$, target item $x_t$ and label $y^*$. The interest hidden states of target item $z_t$ is sampled from $P_t(z)$, and target item $x_t$ are sampled from the conditional distribution $P(x|z)$.
	Interest hidden state sequence of clicked items $\tilde{z}=\{z_1,\cdots,z_p\}$ is sampled from a set $\tilde{S_z}\subset \bigcup_1^{p}Z$. Each element of $\tilde{z}$ controls the user's click behavior in a period of length $T$. 
	That is, items of historical click behaviors subsequence $\{x^{T\times(i-1)+1},\cdots,x^{T\times i}\}_h$ 
	is sampled from the conditional distribution $P(x|z_i)$. The element number of set $S_z$ is $N_{S}$. The label $y^*$ is sampled from $\{0,1\}$.
\end{Assumption}

\begin{Theorem}
	\label{Theorem2}
	If the $N$ training samples are sampled from the distribution under Assumption \ref{Assum2}. For the $D$ layers MLP with ReLU, $f(x_h,x_t)$ and the loss function $l(f,s)=|f(x_h,x_t)-y^*|$, $s=\{\{x^1,\cdots,x^{T\times p}\}_h,x_t,y^*\}$. With the probability $1-\delta$, the difference between expected loss and empirical loss will be \vspace{-0.5cm}
	
	\begin{equation}
	\label{equ2}
	\small{
		\begin{aligned}
		&|E_s(l(f,s))-\frac{1}{N}\sum_{i=1}^{N}l(f,s_i)| \leq \inf_r\{\sqrt{Tp+1}\|W\|_2^Dr\\
		&+ l_M\sqrt{\frac{4N_zN_S(2R_{max}\sqrt{d}/r)^{d(Tp+1)}ln2+2ln(1/\delta)}{N}}\}
		\end{aligned}}
	\end{equation}
	
	Among them, $l_M$ is the maximum value of $l(f,s)$. $d$ is the dimension of each input vector. domain$P_i\in \phi(R_{max})$. $\|W\|_2$ is the average of 2-norm of all parameter matrices.
\end{Theorem}

\begin{Proof}
	According to Assumption \ref{Assum2}, each sample $\{\{x^1,\cdots,x^{T\times p}\}_h,x_t,y\}$ exists a behavior interest hidden state sequence $\tilde{z}_h$ and target interest hidden state $z_t$. 
	
	According to the Assumption \ref{Assum2}, when $\tilde{z}_h=\{z_1,\cdots,z_p\}$ and $z_t=j$ are determined, the domain of elements in each prieod $x_{ph}^i =\{x^{1+T(i-1)},\cdots,x^{T+T(i-1)}\}$ is  domain$P_{z_i}\in\phi(R_{z_i})$, and domain of $x_t$ is domain$P_{j}\in \phi(R_j)$. According to Lemma \ref{lemma}, domain$P_{z_i}$ exists $L_{z_i} = (2R_{z_i}\sqrt{d}/r)^d$ $r$-covering disjoint sets $\{K_{1}^{z_i},\cdots,K_{L_{z_i}}^{z_i}\}$, and domain$P_{j}$ exists $L_j = (2R_{j}\sqrt{d}/r)^d$ $r$-covering disjoint sets $\{K_{1}^j,\cdots,K_{L_j}^j\}$.

	For one sample $s = \{\{x^1,\cdots,x^{T\times p}\}_h,x_t,y^*\}$ in training samples, we define the situation that $\tilde{z}_h = \tilde{z}$, $z_t= j$, each $x^{g}$ in $\{x^1,\cdots,x^{T\times p}\}_h$ and $x^g \in x^i_{ph}$. $x^g \in K^{z_i}_{\alpha}$, $x_t \in K^j_{\beta}$, and $y^* = q \in\{0,1\}$ as $s \in \Psi = \psi (\tilde{z},j,\alpha,\beta,q)$. When $\tilde{z}$ and $j$ remain unchanged, there are $(\prod_{i=1}^{p}(\prod_{1}^{T}L_{z_i}))\times L_j \times 2$ cases in this set. Therefore, The total number of the set $\psi (\tilde{z},j,\alpha,\beta,q)$ is 
	\begin{equation}
	\label{N_2}
	N_{\psi } = \sum_{\tilde{z}\in \tilde{S}_z}\sum_{j=1}^{N_z} 2(\prod_{i=1}^{p}(\prod_{1}^{T}L_{z_i}))L_j\leq 2N_zN_S(2R_{max}\sqrt{d}/r)^{d(Tp+1)}
	\end{equation}
	The difference between expected loss and empirical loss is
	\begin{equation}
	\begin{aligned}
	\label{main_2}
	&|E_s(l(f,s))-\frac{1}{N}\sum_{i=1}^{N}l(f,s_i)|\leq \\
	&|E_s(l(f,s))-\frac{1}{N} \sum_{\Psi}n_{\Psi}E_{s_{\Psi}}(l(f,s_{\Psi})|s_{\Psi}\in \Psi)| +\\
	&|\frac{1}{N} \sum_{\Psi}n_{\Psi}E_{s_{\Psi}}(l(f,s_{\Psi})|s_{\Psi}\in \Psi) -\frac{1}{N}\sum_{i=1}^{N}l(f,s_i)| 
	\end{aligned}
	\end{equation}

	In (\ref{main_2}), we add and subtract one item $\frac{1}{N} \sum_{\Psi}n_{\Psi}E_{s_{\Psi}}(l(f,s_{\Psi})|s_{\Psi}\in \Psi)$. Among them, $n_{\Psi}$ is the number of samples belonging to $\Phi$ in $N$ training samples. $E_{s_{\Psi}}(l(f,s_{\Psi})|s_{\Psi}\in \Psi)$ is the expectation of the loss function when the sample belongs to $\Psi$.
	
	Based on the definition of expectation, $E_s(l(f,s)) = \sum_{\Psi}E_{s_{\Psi}}(l(f,s_{\Psi})|s_{\Psi} \in \Psi )P(s_{\Psi}\in \Psi)$, and $\frac{1}{N}\sum_{i=1}^{N}l(f,s_i) = \frac{1}{N}\sum_{\Psi}\sum_{s_i \in \Psi} l(f,s_i)$. The equation (\ref{main_2}) will be 
	\begin{equation}
	\begin{aligned}
	\label{main2_2}
	&|E_s(l(f,s))-\frac{1}{N}\sum_{i=1}^{N}l(f,s_i)| \leq \\
	&\sum_{\Psi}|E_{s_{\Psi}}(l(f,s_i)|s_{\Psi} \in \Psi )||P(s_{\Psi}\in \Psi) - \frac{n_{\Psi}}{N}|+\\
	&\frac{1}{N}\sum_{\Psi}|n_{\Psi}E_{s_{\Psi}}(l(f,s_{\Psi})|s_{\Psi} \in \Psi )-\sum_{s_i \in \Psi}l(f,s_i)|\leq \\
	&l_M\sum_{\Psi}|P(s_i\in \Psi) - \frac{n_{\Psi}}{N}| + \frac{1}{N}\sum_{\Psi}\sum_{s_i \in \Psi}|E_{s_{\Psi}}(l(f,s_{\Psi})|s_{\Psi} \in \Psi)-l(f,s_i)|
	\end{aligned}
	\end{equation}
	$l_M$ is the maximal value of $l(f,s)$.
	According to the concentration inequalities for multinomial distribution in Lemma \ref{lemma3}, in this situation, $K = N_{\Psi}$. Therefore,
	$P(\sum_{\Psi}|P(s_i\in \Psi) - \frac{n_{\Psi}}{N}| \leq \lambda) \geq 1 - 2^{N_{\Psi}}\exp(\frac{-n\lambda^2}{2})$. Let $\delta = 2^{N_{\Psi}}\exp(\frac{-n\lambda^2}{2})$, then $\lambda = \sqrt{\frac{2N_{\Psi}ln2+2ln(1/\delta)}{2}}$. It means that 
	\begin{equation}
	\label{concen_2}
	\sum_{\Psi}|P(s_i\in \Psi) - \frac{n_{\Psi}}{N}| \leq \sqrt{\frac{2N_{\Psi}ln2+2ln(1/\delta)}{n}}
	\end{equation}
	with at least probability $1-\delta$.

	$\forall a,s \in \Psi$, $a =\{\{x^1,\cdots,x^{Tp}\}_h^1,x_t^1,y^1\}$,$s =\{\{x^1,\cdots,x^{Tp}\}_h^2,x_t^2,y^2\}$. elements of $\{x^1,\cdots,x^{Tp}\}_h^1$ and $\{x^1,\cdots,x^{Tp}\}_h^2$ belong to the same $r$-covering disjoint set. This is the same as the case in Proof \ref{Proof1}, the difference in the individual component of the input is not more than $r$. The $n$ in Lemma \ref{lemma2} is $T\times p+1$. Therefore, $\|x^1-x^2\|_2\leq \sqrt{Tp+1}r$. According to Lemma \ref{lemma2}, there will be $|l(f,a)-l(f,s)|\leq \sqrt{Tp+1}\|W\|_2^Dr$,
	\begin{equation}
	|E_{s_{\Psi}}(l(f,s_{\Psi})|s_{\Psi} \in \Psi)- l(f,s_i)|\leq \sqrt{Tp+1}\|W\|_2^Dr.
		\end{equation}
	We substitute it in (\ref{main2_2}) as 
	\begin{equation}
	\begin{aligned}
	&|E_s(l(f,s))-\frac{1}{N}\sum_{i=1}^{N}l(f,s_i)| \leq \\
	&\sqrt{\frac{2N_{\Psi}ln2+2ln(1/\delta)}{n}} + \sqrt{Tp+1}\|W\|_2^Dr.
	\end{aligned}
	\end{equation}
	In addition, $r$ is the dividing radius of the input domain, and it can be adjusted arbitrarily. Therefore, the generalization error bound should be smaller than infimum under all $r$. According to (\ref{N_2}), (\ref{concen_2}),with at least probability $1-\delta$
	\begin{equation}
	\begin{aligned}
	&|E_s(l(f,s))-\frac{1}{N}\sum_{i=1}^{N}l(f,s_i)| \leq \\
	&\inf_r\{\sqrt{\frac{4N_zN_S(2R_{max}\sqrt{d}/r)^{d(Tp+1)}ln2+2ln(1/\delta)}{N}} + \sqrt{Tp+1}\|W\|_2^Dr\}.
	\end{aligned}
	\end{equation}
	
\end{Proof}

\section{Consistency between method and theory}
According to the modeling of the user behavior, the user's interest will remain unchanged in a period of time, and the domain of interest of the clicked product will remain unchanged in this period of time. It means that two items are more likely to be in the same interest domain if they co-occur more frequently in a short-term of user historical behaviors. In summary, interconnected items in the graph are more likely to be in the same interest domain.

In the actual scenario, the relationship of interest domains among items is quite complex(e.g. a item may belong to several interest domains) so that the connection in interest graph is complex. For convenience of analysis, we consider a special case, that is, each item belongs to a single interest domain. In this case, each interest domain becomes a isolated island in interest graph. The items in one isolated island are interconnected, and the items between different isolated islands are not connected. In this section, we will ultilize this case to illustrate that the average operation on central embedding basis matrix will be consistent with the conclusion of theoretical analysis in Proposition \ref{propostion 1}, that is, our proposed methods could reduce the distance among items with the same interest in the embedding space and maintain the distance of items with different interest domains in embedding space.

\begin{Proposition}
	\label{propostion 1}
	Assume there are $N_l$ isolated islands in the item interest graph $\bm{Z}$, the number of items of island $i$ is $m_i$. The items in the isolated island are connected to each other, and the items in different islands are not connected to each other in the graph. The embedding matrix is $\bm{X}$ and embedding vector of $q$-th item is $\bm{X}(q,:)$. The center of the island $i$ is $c(\bm{X},i)=\sum_{q\in i}\bm{X}(q,:)/m_i$. The internal distance of island $i$ is $m_s (\bm{X},i)=\sum_{q\in i}\|\bm{X}(q,:)- c(\bm{X},i)\|_2/m_i$, and the average distance between islands is $m (\bm{X}, i, j) = \|c(\bm{X},i)-c(\bm{X},j)\|_2$. Through the operation $g_{AVG}()$ defined in (6) in the main paper, the new embedding matrix $\bm{X}' = g_{AVG}(\bm{Z})\bm{X}$. The internal distance of island $i$ becomes $m_s(\bm{X}',i) = m_s(\bm{X},i)/(m_i-1)$. The average distance between two islanded islands remains unchanged, $m (\bm{X}', i, j)  = m (\bm{X}, i, j) $.
\end{Proposition}
The conclusion of Proposition \ref{propostion 1} tells us, after the operation on central embedding basis matrix proposed in (6) in the main paper, $\bm{X}' = g_{AVG}(Z)\bm{X}$, the internal distance of each island shrunk to $\frac{1}{m_i-1}$ times. Correspondingly, as each isolated island in the graph is corresponding to an interest domain, the envelope radius of embedding vectors of items with the same interest domain also shrinks to $\frac{1}{m_i-1}$ times. In E-commerce scenarios, items with similar categories is often very sufficient so that number of items in each interest domain $m_i$ is quite large. If we maintain the largest $K$ connection weights of each item in the interest graph and cut rest of them like Algorithm 1 in the paper, then $m_i = K+1$. Therefore, $\frac{1}{m_i-1}<1$ or even $\frac{1}{m_i-1}<<1$ i.e $R_{max}$ in (\ref{equ1}) and (\ref{equ2}) could be greatly reduced through operation $\bm{X}' = g_{AVG}(Z)\bm{X}$. At the same time, the average distances among islands are constant after operation on central embedding basis matrix. In summary, using the operation on central embedding of res-embedding help reduce the distance of items in the same interest domain in embedding space and keep the distance of items with different interest domain in embedding space unchanged. This conlusion demonstrate that res-embedding could control the generalization error bound of the CTR model effectively. The proof of Proposition \ref{propostion 1} is deduced in the supplymental material.

\section{Proof of Proposition 1}

\begin{Proposition}
	\label{propostion 1}
	Assume there are $N_l$ isolated islands in the item interest graph $\bm{Z}$, the number of items of island $i$ is $m_i$. The items in the isolated island are connected to each other, and the items in different islands are not connected to each other in the graph. The embedding matrix is $\bm{X}$ and embedding vector of $q$-th item is $\bm{X}(q,:)$. The center of the island $i$ is $c(\bm{X},i)=\sum_{q\in i}\bm{X}(q,:)/m_i$. The internal distance of island $i$ is $m_s (\bm{X},i)=\sum_{q\in i}\|\bm{X}(q,:)- c(\bm{X},i)\|_2/m_i$, and the average distance between islands is $m (\bm{X}, i, j) = \|c(\bm{X},i)-c(\bm{X},j)\|_2$. Through the operation $g_{AVG}()$ defined in (8)(At the main paper.) , the new embedding matrix $\bm{X}' = g_{AVG}(\bm{Z})\bm{X}$. The internal distance of island $i$ becomes $m_s(\bm{X}',i) = m_s(\bm{X},i)/(m_i-1)$. The average distance between two islanded islands remains unchanged, $m (\bm{X}', i, j)  = m (\bm{X}, i, j) $.
\end{Proposition}

\begin{Proof}
	For an isolate island $i$, $\forall q\in i$, there will be
	\begin{equation}
	\label{1}
	\bm{X'}(q,:)=g_{AVG}(\bm{Z})(q,:)\bm{X}.
	\end{equation}
	 According to the definition of $g_{AVG}()$.
	\begin{equation}
		\label{2}
	g_{AVG}(\bm{Z}) = {\rm avg_{nz}}(I(\bm{Z} > 0)).
	\end{equation}
	 The indicator function $I (\bm{Z} > 0)$ means that each element of the $F$ is changed to 1 if it is greater than 0, otherwise changed to 0.
	The ${\rm avg_{nz}}$ operation averages all the non-zero elements of every row vector in the matrix, and the 0 elements remain unchanged.
	
	Because of $Z(q,p)>0$ for $\forall p\in i$ there will be $g_{AVG}(\bm{Z})(q,p)=1/(m_i-1)$. Correspondingly, $\forall p \in i$ and $g_{AVG}(\bm{Z})(q,p)=0$, for $\forall p \notin i$. According to (\ref{1}) and (\ref{2}),
	
	\begin{equation}
	\bm{X'}(q,:)=\sum_{p \in i,p\neq q}\frac{\bm{X}(p,:)}{m_i-1}.
	\end{equation}
	The new center of the island $i$, 
	\begin{equation}
	\begin{aligned}
	\label{same_central}
	c(\bm{X}',i)&=\sum_{q\in i}\bm{X}'(q,:)/m_i\\
	&=\sum_{q \in i}\sum_{p \in i,p \neq q}\bm{X}(p,:)/(m_i)(m_i-1)\\
	&=\sum_{q \in i}(\sum_{p \in i}\bm{X}(p,:)-\bm{X}(q,:))/(m_i)(m_i-1)\\
	&=(m_i-1)(\sum_{q \in i}\bm{X}(q,:)/(m_i)(m_i-1))\\
	&=(\sum_{q \in i}\bm{X}(q,:)/(m_i))=c(\bm{X},i)
	\end{aligned}
	\end{equation}
	(\ref{same_central}) illustrates that when the embedding matrix $\bm{X}$ transforms from linear to X', the center of each island remains unchanged.
	\begin{equation}
	\begin{aligned}
	m(\bm{X}',i,j) &= \|c(\bm{X}',i)-c(\bm{X}',j)\|_2\\
	&= \|c(\bm{X},i)-c(\bm{X},j)\|_2\\
	&= m(\bm{X},i,j) 
	\end{aligned}
	\end{equation}
	The second conclusion is proved.
	
	The internal distance of island $i$ based on the $\bm{X}'$ is
		\begin{equation}
	\begin{aligned}
	m_s(\bm{X}',i)&=\sum_{p \in i}\|\bm{X}'(q,:)-c(\bm{X}',i)\|_2/m_i\\
	&=\sum_{q \in i}\|\sum_{p \in i,p\neq q}\frac{\bm{X}(p,:)}{m_i-1}-\sum_{q \in i}\bm{X}(q,:)/(m_i)\|_2/m_i\\
	&=\sum_{q \in i}\|\frac{\sum_{p \in i}\bm{X}(p,:)-\bm{X}(q,:)}{m_i-1}-\sum_{q \in i}\bm{X}(q,:)/(m_i)\|_2/m_i\\
	&=\sum_{q \in i}\|\frac{m_i\sum_{p \in i}\bm{X}(p,:)-m_i\bm{X}(q,:)-(m_i-1)(\sum_{q \in i}\bm{X}(q,:))}{(m_i-1)m_i}\|_2/m_i\\
	&=\sum_{q \in i}\|\frac{\sum_{p \in i}\bm{X}(p,:)-m_i\bm{X}(q,:)}{(m_i-1)m_i}\|_2/m_i\\
	&=\sum_{q \in i}\frac{\|\sum_{q \in i}\bm{X}(p,:)/m_i-\bm{X}(q,:)\|_2}{m_i(m_i-1)}\\
	&=\sum_{q \in i}\frac{\|c(\bm{X},i)-\bm{X}(q,:)\|_2}{m_i}/(m_i-1)=m_s(\bm{X},i)/(m_i-1)\\
	\end{aligned}
	\end{equation}
	The both conclusions of Proposition \ref{propostion 1} are proved.
\end{Proof}
\bibliography{appendix}